# Building and Refining Abstract Planning Cases
# by Change of Representation Language

**Ralph Bergmann**                                     BERGMANN@INFORMATIK.UNI-KL.DE
**Wolfgang Wilke**                                        WILKE@INFORMATIK.UNI-KL.DE
*Centre for Learning Systems and Applications (LSA)*
*University of Kaiserslautern, P.O.-Box 3049, D-67653 Kaiserslautern, Germany*

## Abstract

Abstraction is one of the most promising approaches to improve the performance of problem solvers. In several domains abstraction by *dropping sentences* of a domain description – as used in most hierarchical planners – has proven useful. In this paper we present examples which illustrate significant drawbacks of abstraction by dropping sentences. To overcome these drawbacks, we propose a more general view of abstraction involving the *change of representation language*. We have developed a new abstraction methodology and a related sound and complete learning algorithm that allows the complete *change of representation language* of *planning cases* from concrete to abstract. However, to achieve a powerful change of the representation language, the abstract language itself as well as rules which describe admissible ways of abstracting states must be provided in the domain model. This new abstraction approach is the core of PARIS (**P**lan **A**bstraction and **R**efinement in an **I**ntegrated **S**ystem), a system in which abstract planning cases are automatically learned from given concrete cases. An empirical study in the domain of process planning in mechanical engineering shows significant advantages of the proposed reasoning from abstract cases over classical hierarchical planning.

## 1. Introduction

*Abstraction* is one of the most challenging and also promising approaches to improve complex problem solving and it is inspired by the way humans seem to solve problems. At first, less relevant details of a given problem are ignored so that the abstracted problem can be solved more easily. Then, step by step, more details are added to the solution by taking an increasingly more detailed look at the problem. Thereby, the abstract solution constructed first is refined towards a concrete solution. One typical characteristic of most work on hierarchical problem solving is that abstraction is mostly performed by *dropping sentences* of a domain description (Sacerdoti, 1974, 1977; Tenenberg, 1988; Unruh & Rosenbloom, 1989; Yang & Tenenberg, 1990; Knoblock, 1989, 1994; Bacchus & Yang, 1994). A second common characteristic is that a hierarchical problem solver usually derives an abstract solution from scratch, without using experience from previous problem solving episodes.

Giunchiglia and Walsh (1992) have presented a comprehensive formal framework for abstraction and a comparison of the different abstraction approaches from theorem proving (Plaisted, 1981, 1986; Tenenberg, 1987), planning (Newell & Simon, 1972; Sacerdoti, 1974, 1977; Tenenberg, 1988; Unruh & Rosenbloom, 1989; Yang & Tenenberg, 1990; Knoblock, 1989, 1994), and model based diagnosis (Mozetic, 1990). For hierarchical planning, Korf's model of abstraction in problem solving (Korf, 1987) allows the analysis of reductions in search caused by single and multiple levels of abstraction. He has shown that in the optimal case, abstraction can reduce the expected search time from exponential to linear. Knoblock has developed an approach to construct a hierarchy of abstraction spaces automatically from a given concrete-level problem solving domain (Knoblock, 1990, 1993, 1994). These so called ordered monotonic abstraction hierarchies (Knoblock, Tenenberg, & Yang, 1991b) have proven useful in many domains. Recently, Bacchus and Yang (1994) presented an improved method for automatically generating abstraction hierarchies based on a more detailed model of search costs.

All these abstraction methods, however, rely on abstraction by dropping sentences of the domain description which is a kind of *homomorphic abstraction* (Holte et al., 1994, 1995). It has been shown that these kinds of abstractions are highly representation dependent (Holte et al., 1994, 1995). For two classical planning domains, different "natural" representations have been analyzed and it turns out that there are several representations for which the classical abstraction techniques do not lead to significantly improved problem solvers (Knoblock, 1994; Holte et al., 1995). However, it is well known that normally many different representations of the same domain exist as already pointed out by Korf (1980), but up to now no theory of representation has been developed. In particular, there is no theory of representation for hierarchical problem solving with dropping sentences.

From a knowledge-engineering perspective, many different aspects such as simplicity, understandability, and maintainability must be considered when developing a domain representation. Therefore, we assume that representations of domains are given by knowledge engineers and rely on representations which we consider most "natural" for certain kinds of problems. We will demonstrate two simple example problems and related representations, in which the usual use of abstraction in problem solving does not lead to any improvement. In the first example, no improvement can be achieved because abstraction is restricted to dropping sentences of a domain. In the second example, the abstract solution computed from scratch does not decompose the original problem and consequently does not cut down the search space at the next detailed level. We do not want to argue that the examples can never be represented in a way that standard hierarchical problem solving works well. However, we think it would require a large effort from a knowledge engineer to develop an appropriate representation and we believe that it is often impossible to develop a representation which is appropriate from a knowledge-engineering perspective and which also allows efficient hierarchical problem solving based on dropping sentences.

We take these observations as the motivation to develop a more general model of abstraction in problem solving. As already pointed out by Michalski (1994), abstraction, in general, can be seen as switching to a completely new representation language in which the level of detail is reduced. In problem solving, such a new abstract representation language must consist of completely new sentences and operators and not only of a subset of the sentences and operators of the concrete language. To our knowledge, SIPE (Wilkins, 1988) is the only planning system which currently allows the change of representation language across different levels of abstraction. However, a general abstraction methodology which allows efficient algorithms for abstraction and refinement has not yet been developed. We want to propose a method of abstraction which allows the *complete change of representation language* of a problem and a solution from concrete to abstract and vice versa, if the concrete and the abstract language are given. Additionally, we propose to use *experience* from previously solved problems, usually available as a set of *cases*, to come to abstract solutions. The use of experience has already proven useful in various approaches to speed-up learning such as explanation-based learning (Mitchell, Keller, & Kedar-Cabelli, 1986; DeJong & Mooney, 1986; Rosenbloom & Laird, 1986; Minton, 1988; Minton, Carbonell, Knoblock, Kuokka, Etzioni, & Gil, 1989; Shavlik & O'Rorke, 1993; Etzioni, 1993; Minton & Zweben, 1993; Langley & Allen, 1993; Kambhampati & Kedar, 1994), and analogical or case-based reasoning (Carbonell, 1986; Kambhampati & Hendler, 1992; Veloso & Carbonell, 1993; Veloso, 1994).

As the main contribution of this paper, we present an abstraction methodology and a related *learning* method in which beneficial *abstract planning cases* are automatically derived from given concrete cases. Based on a given *concrete and abstract language*, this learning approach allows the complete change of the representation of a case from the concrete to the abstract level. However, to achieve such an unconstrained kind of abstraction, the set of admissible abstractions must be implicitly predefined by a *generic abstraction theory*. Compared to approaches in which abstraction hierarchies are generated automatically, more effort is required to specify the abstract language, but we feel that this is a price we have to pay to make planning more tractable in certain situations.

This approach is fully implemented in PARIS (**P**lan **A**bstraction and **R**efinement in an **I**ntegrated **S**ystem), a system in which abstract cases are learned and organized in a *case base*. During novel problem solving, this case-base is searched for a suitable abstract case which is further *refined* to a concrete solution to the current problem.

The presentation of this approach is organized as follows. The next section presents an analysis of hierarchical problem solving in which the shortcomings of current approaches are illustrated by simple examples. Section three argues that a powerful case abstraction and refinement method can overcome the identified problems. Furthermore, we present the PARIS approach informally, using a simple example. The next three sections of the paper formalize the general abstraction approach. After introducing the basic terminology, Section 5 defines a new formal model of case abstraction. Section 6 contains a very detailed description of a correct and complete learning algorithm for case abstraction. Section 7 explains the refinement of cases for solving new problems. Section 8 gives a detailed description of the domain of process planning in mechanical engineering for the production of rotary-symmetric workpieces on a lathe and demonstrates the proposed approach on examples from this domain. Section 9 reports on a detailed experimental evaluation of PARIS in the described domain. Finally, we discuss the presented approach in relation to similar work in the field. The appendix of the article contains the formal proofs of the properties of the abstraction approach and the related learning algorithm. Additionally, the detailed representation of the mechanical engineering domain used for the experimental evaluation is given in Online Appendix 1.

## 2. Analysis of Hierarchical Problem Solving

The basic intuition behind abstraction is as follows. By first ignoring less relevant features of the problem description, abstraction allows problems to be solved in a coarse fashion with less effort. Then, the derived abstract (skeletal) solution serves as a problem decomposition for the original, more detailed problem. Korf (1987) has shown that hierarchical problem solving can reduce the required search space significantly. Assume that a problem requires a solution of length $n$ and furthermore assume that the average branching factor is $b$, i.e., the average number of states that can be reached from a given state by applying a single operator. The worst-case time complexity for finding the required solution by search is $O(b^n)$. Now, suppose that the problem is decomposed by an abstract solution into $k$ subproblems, each of which require a solution of length $n_1, \ldots, n_k$, respectively, with $n_1 + n_2 + \cdots + n_k = n$. In this situation, the worst-case time complexity for finding the complete solution is $O(b^{n_1} + b^{n_2} + \cdots + b^{n_k})$ which is $O(b^{max(n_1, n_2, \ldots, n_k)})$. Please note that this a significant reduction in search time complexity. In particular, we can easily see that the reduction is maximal if all subproblems are of similar size, i.e., $n_1 \approx n_2 \approx \cdots \approx n_k$.

However, to achieve a significant search reduction, the computed abstract solution must not only be a solution to the abstracted problem, it must additionally fulfill a certain requirement presupposed in the above analysis. The subproblems introduced by the abstract solution must be independent, i.e., each of them must be solvable without interaction with the other subproblems. This avoids backtracking between the solution of each subproblem and consequently cuts down the necessary overall search space. Even if this restriction is not completely fulfilled, i.e., backtracking is still required in a few cases, several empirical studies (especially Knoblock, 1991, 1993, 1994) have shown that abstraction can nevertheless lead to performance improvements.

Unfortunately, there are also domains and representations of domains (Holte et al., 1994, 1995) in which the way abstraction is used in hierarchical problem solving cannot improve problem solving because the derived abstract solutions don't fulfill the above mentioned requirement at all. In the following, we will show two examples of such domains which demonstrate two general drawbacks of hierarchical problem solving. Please note that in these examples, a particular representation is assumed. We feel that these representations are somehow "natural" and very likely to be used by a knowledge engineer developing a domain. However, there might be other representations of these domains for which traditional hierarchical planning works. We assume that such representations are very difficult to find, especially if the domain representation should also fulfill additional knowledge-engineering requirements.

## 2.1 Abstraction by Dropping Sentences

In hierarchical problem solving, abstraction is mostly[1] achieved by *dropping sentences* of the problem description from preconditions and/or effects of operators (Sacerdoti, 1974, 1977; Tenenberg, 1988; Unruh & Rosenbloom, 1989; Yang & Tenenberg, 1990; Knoblock, 1989, 1994). The assumption which justifies this kind of abstraction is that less relevant details of the problem description are expressed as isolated sentences in the representation which can be addressed after the more relevant sentences have been established. Ignoring such sentences is assumed to lead to an abstract solution useful to reduce the search at the more concrete planning levels.

However, this assumption does not hold in all domains. For example, in many real world domains, certain events need to be counted, e.g., when transporting a certain number of containers from one location to another. Imagine a domain in which, in addition to several other operators, there is an *increment operator* described as follows:

> *Operator:* `inc`
> *Precondition:* `value(X)`
> *Delete:* `value(X)`
> *Add:* `value(X + 1)`

---

1. Only Tenenberg's (1988) abstraction by analogical mappings and the planning system SIPE (Wilkins, 1988) contains first approaches that allow a change of representation language.

In this representation, the integer value which is increased is represented by a single sentence. Each state consists only of a single sentence, and also the operator contains only one single sentence.[2] We think that this representation is very "natural" and very likely to be chosen by a knowledge engineer. In this domain, incrementing `value(0)` to `value(8)` requires a sequential plan composed of 8 `inc`-operators, leading to the state sequence: `value(0)`,`value(1)`,...,`value(8)`. In this example, however, abstraction by dropping sentences does not work because, if this single sentence would be dropped, nothing would remain in the operator description and the whole counting problem would have been dropped completely. So there is only the empty problem at the abstract level, and the empty plan is going to solve it. Unfortunately, the empty plan cannot cause any complexity reduction for solving the problem at the concrete level. Consequently, abstraction by dropping sentences completely fails to improve problem solving in this situation.

However, we can adequately cope with this counting problem by abstracting the quantitative value expressed in the sentence towards a qualitative representation (e.g., `low`=$\{0, 1, 2, 3\}$, `medium` = $\{4, 5, 6, 7\}$, `high` = $\{8, 9, 10, 11\}$). Such a qualitative representation would result in an abstract plan composed of two operators (subproblems) that increase `value` from `low` to `medium` and further to `high`. This abstract plan defines two independently refinable subproblems. To solve the first subproblem at the concrete level, the problem solver has to search for a sequence of `inc`-operators which increment the value from 0 to any medium value (any value from the set $\{4, 5, 6, 7\}$). This subproblem can be solved by a sequence of 4 `inc`-operators leading to the concrete state with a value of 4. Similarly, the second subproblem at the concrete level is to find a sequence of operators which change the value from 4 to the final value 8. Also this second subproblem can be solved by a sequence of 4 `inc`-operators. So we can see that the complete problem which requires a sequence of 8 concrete operators is divided into 2 subproblems where each subproblem can be solved by a 4-step plan. Because of the exponential nature of the search space, the two 4-step problems together can be solved with much less search than the 8-step problem as a whole. Following Korf's analysis sketched before, the time complexity is reduced from $O(b^8)$ to $O(b^4)$[3]. Please note that the particular abstraction which leads to two subproblems is not central for achieving the complexity reduction. The important point is that the problem is decomposed into more than one subproblem. This kind of abstraction can be achieved by introducing a new abstract representation language which consists of the qualitative values and a corresponding abstract increment operator.

---

2. However, we might assume that the term $X + 1$ is modeled as a separate predicate in the precondition. Unfortunately, this does not change the described situation at all.

3. Because we assume many other operators besides the `inc`-operator, $b \gg 1$ holds.

We can even generalize from the specific example presented above. The problem with the dropping condition approach is that it is not possible to abstract information (e.g., the value in our example) that is coded in a single sentence in the representation. This is particularly a problem when the required solution contains a long sequence of states which only differ in a single sentence. Dropping this particular sentence leads to dropping the whole problem, and not dropping the sentence does not lead to any abstraction. What is really required is to abstract the information encoded in this single sentence which obviously requires more than just dropping the complete information.

To summarize, we have seen that abstraction by dropping sentences does not work for the particular kind of problems we have shown. In general, abstraction requires changing the *complete representation language* from concrete to abstract which usually involves the introduction of completely new abstract terms (sentences or operators). Within this general view, dropping sentences is just a special case of abstraction. The reason why dropping sentences has been widely used in hierarchical planning is that due to its simplicity, refinement is very easy because abstract states can directly be used as goals at the more detailed levels. Another very important property of abstraction by dropping sentences is that useful hierarchies of abstraction spaces can be constructed automatically from domain descriptions (Knoblock, 1990, 1993, 1994; Bacchus & Yang, 1994).

## 2.2 Generating of Abstract Solutions from Scratch

Another limiting factor of classical hierarchical problem solving can be the way abstract solutions are computed. As pointed out by Korf, a good abstract solution must lead to mostly independent subproblems of equal size. In classical problem solving, an abstract solution is found by breadth-first or depth-first search using linear (e.g., ALPINE, Knoblock, 1993) or non-linear (e.g., ABTWEAK, Yang & Tenenberg, 1990) problem solvers. For these problem solvers, the upward-solution property (Tenenberg, 1988) usually holds, which means that an abstract solution exists if a concrete-level solution exists. Usually, these problem solvers find an *arbitrary* abstract solution (e.g., the shortest possible solution). Unfortunately, there is no way to guarantee that the computed solutions are refinable and lead to mostly independent subproblems of sufficiently equal size, even if such a solution exists. In general, there are not even heuristics which try to guide problem solving towards the aspired kind of useful abstractions. This problem is illustrated by the following example, which additionally shows the limitation of abstraction by dropping sentences.

Imagine a large (or even infinite) state space which includes at least the 8 distinct states shown on the left of Figure 1. Each of these 8 states is described by the presence or absence of three sentences $E1$, $E2$, and $E3$ in the state description. In the 3-bit-vector shown in Figure 1, "0" indicates the absence of the sentence and "1" represents the presence of the sentence. The 8 different states described by these three sentences are arranged in a 3-dimensional cube, using one dimension for each sentence. The arrows in this diagram show possible state transitions by the available operators of the domain.[4] Each operator manipulates (adds or deletes) exactly one sentence of the state description, if certain conditions on the other sentences are fulfilled. The representation of two of these operators is shown on the right

---

4. The dashed lines do not represent operators and are only introduced to make the shape of the cube more easy to see.

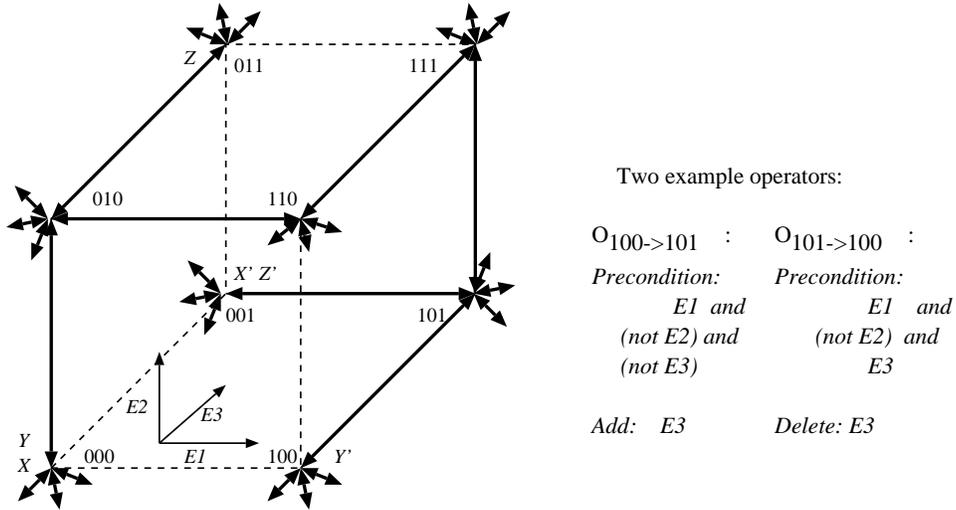

Two example operators:

$O_{100\to101}$ :          $O_{101\to100}$ :

*Precondition:*          *Precondition:*
    *E1 and*              *E1 and*
    *(not E2) and*              *(not E2) and*
    *(not E3)*              *E3*

*Add:  E3*          *Delete: E3*

Figure 1: State space of an example domain and representation of two operators

side of this figure. The subscript of the operator name relates to the respective transition in the state diagram. In general, we can see that

- $E1$ can be manipulated, if and only if $E2 \oplus E3$ holds,

- $E2$ can be manipulated, if and only if $E1 \equiv E3$ holds, and

- $E3$ can be manipulated, if and only if $E1 \lor E2$ holds.

Furthermore, assume that there are many more operators which connect some other states of the domain, not shown in the diagram, to the 8 depicted states. Consequently, we must assume a branching factor of $b \gg 1$ at each state, which makes the search space for problem solving quite large. Besides the description of the domain, Figure 1 also shows three example problems: $X \to X'$, $Y \to Y'$ and $Z \to Z'$. For example, the solution to the problem $X \to X'$ is the 5-step path $000 \to 010 \to 110 \to 111 \to 101 \to 001$.

Now, let's consider the abstract solutions which correspond to the concrete solutions for each of the three problems. For each problem, we want to examine the three possible ways of abstraction by dropping one of the sentences. For this purpose, the geometric arrangement of the states turns out to be very useful because abstraction can be simply viewed as projecting the 3-dimensional state space onto the plane defined by the sentences which are not dropped by abstraction. The left part of Figure 2 shows the three possible abstract state spaces which result from dropping one of the sentences. Here it is very important to see that in each abstract state space, every sentence can be modified unconditionally and independent of the other sentences. However, only one sentence can be modified by each operator. Thereby, all the constraints that exist at the concrete level are relaxed. The abstraction of the concrete solution to each of the three problems ($X \to X'$, $Y \to Y'$ and $Z \to Z'$) with respect to the three possible ways of dropping conditions is shown on the right side of Figure 2. Each of the nine possible abstract solutions consists of three or four abstract operators. The sequence in which they have to be applied is indicated by the numbers which mark these operators. We can also see that whatever sentence we drop for any of the problems, an appropriate abstract solution exists which decomposes the original problem into independent refinable subproblems of sufficiently equal size. The main point about this example is that none of these abstract solutions will be found by a hierarchical problem solver! The reason for this is that for each of the abstracted problems there also exists a 0-step or a 1-step solution in addition to the nine 3-step or 4-step solutions indicated by the depicted paths. However, such a short solution is completely useless for reducing the search at the next more concrete level because the original problem is not decomposed at all. The central problem with this is that most problem solvers will find the shorter but useless solutions first, and try to refine them. Consequently, the search space on the concrete level is not reduced so that no performance improvement is achieved at all. However, there might be other representations of this example domain in which a hierarchical problem solver comes to a useful abstract solution. We think, however, that the representation shown is quite natural because it represents the 8 different states with the minimal number of binary sentences.

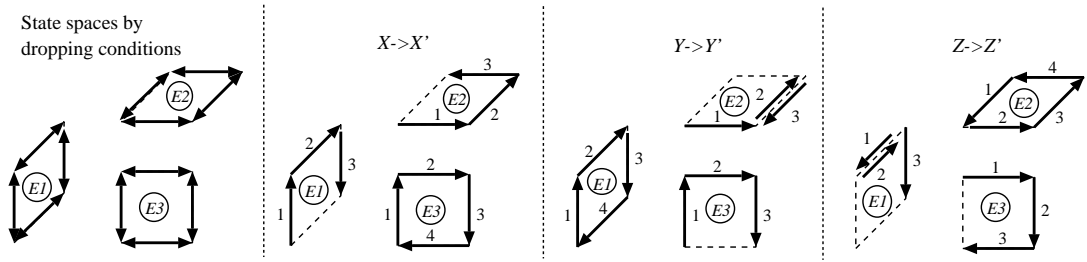

Figure 2: Abstract state spaces by dropping conditions

To summarize, we presented an example in which a useful abstract solution is not found by hierarchical planning although it exists. The reason for this is that planners usually try to find shortest solutions, which is a good strategy for the ground level, but which may not be appropriate at the abstract level. Neither it is desirable to search for the longest solutions because this might cause unnecessarily long concrete plans.

## 3. Case Abstraction and Refinement

As a way out of this problem, we propose to use *experience* given in the form of *concrete planning cases* and to abstract this experience for its reuse in new situations. Therefore, we need a powerful abstraction methodology which allows the introduction of a completely new abstract terminology at the abstract level. This makes it possible that useful abstract solutions can be expressed for domains in which abstraction by dropping conditions is not sufficient. In particular, this methodology must not only serve as a means to analyze different abstraction approaches, but it must allow efficient algorithms for abstracting and refining problems and solutions.

## 3.1 The Basic Idea

We now introduce an approach which achieves case abstraction and refinement by changing the representation language. As a prerequisite, this approach requires that the abstract language itself (state description and operators) is given by a domain expert in addition to the concrete level description. We also require that a set of admissible ways of abstracting states is implicitly predefined by a generic abstraction theory. This is of course an additional knowledge engineering requirement, but we feel that this is a price we have to pay to enhance the power of hierarchical problem solving. Recent research on knowledge acquisition already describes approaches and tools for the acquisition of concrete level and abstract level operators in real-world domains (Schmidt & Zickwolff, 1992; Schmidt, 1994). An abstract language which is given by the user has the additional advantage that abstracted cases are expressed in a language with which the user is familiar. Consequently, understandability and explainability, which are always important issues when applying a system, can be achieved more easily.

As a source for learning, we assume a set of concrete planning cases, each of which consists of a problem statement together with a related solution. As is the case in Prodigy (Minton et al., 1989), we only consider sequential plans, i.e., plans with totally ordered operators. The planning cases we assume do not include a problem solving trace as for example the *problem solving cases* in Prodigy/Analogy (Veloso, 1992; Veloso & Carbonell, 1993; Veloso, 1994). In real-world applications, a domain expert's solutions to previous problems are usually recorded in a company's filing cabinet or database. These cases can be seen as a collection of the company's experience, from which we want to draw power.

During a *learning phase*, a set of abstract planning cases is generated from each available concrete case. An abstract planning case consists of an abstracted problem description together with an abstracted solution. The case abstraction procedure guarantees that the abstract solution contained in an abstract case can always be refined to become a solution of the concrete problem contained in the concrete case that became abstracted. Different abstract cases may be situated at different levels of abstraction or may be abstractions according to different abstraction aspects. Different abstract cases can be of different utility and can reduce the search space at the concrete level in different ways. It can also happen that several concrete cases share the same abstraction. The set of all abstract planning cases that are learned is organized in a case-base for efficient retrieval during problem solving.

During the *problem solving phase*, this case base is searched until an abstract case is found which can be applied to the current problem in hand. An abstract case is applicable to the current problem if the abstracted problem contained in the abstract planning case is an abstraction of the current problem. However, we cannot guarantee that an abstract solution contained in a selected abstract case can really be refined to become a solution to the current problem. It is at least known that each abstract solution from the case base was already useful for solving one or more previous problems, i.e., the problems contained in those concrete cases from which the abstract case was learned. Since the new problem is similar to these previous problems because both can be abstracted in the same way, there is at least a high chance that the abstract solution is also useful for solving the new problem. When the new problem is solved by refinement a new concrete case arises which can be used for further learning.

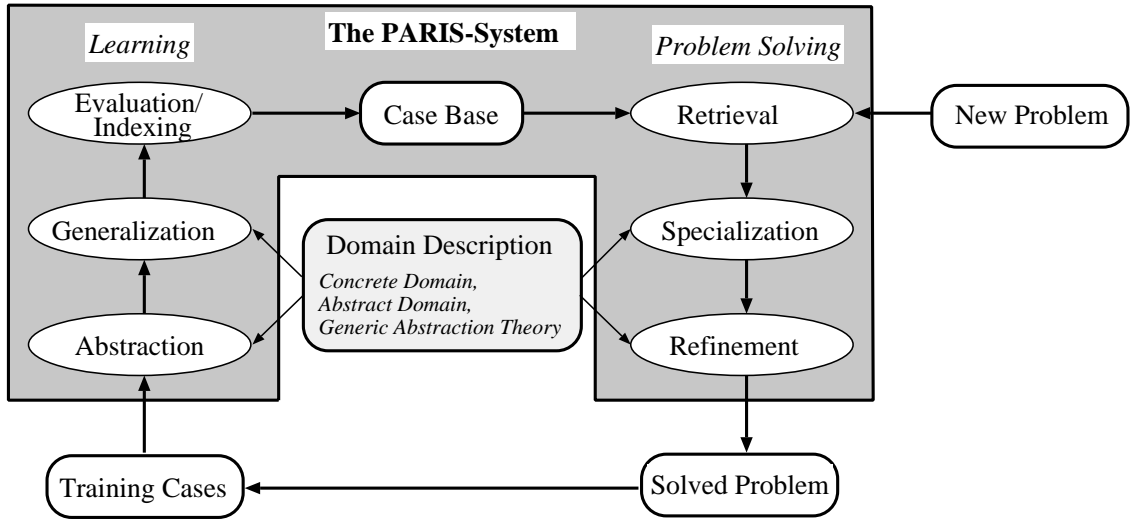

Figure 3: The components of the PARIS System

## 3.2 The PARIS Architecture

PARIS (**P**lan **A**bstraction and **R**efinement in an **I**ntegrated **S**ystem) follows the basic approach just described. Figure 3 shows an overview of the whole system and its components. Besides case abstraction and refinement, PARIS also includes an explanation-based approach for generalizing cases during learning and for specializing them during problem solving. Furthermore, the system also includes additional mechanisms for evaluating different abstract cases and generalizations derived by the explanation-based component. This evaluation component measures the reduction in search time caused by each abstract plan when solving those concrete problems from the case base for which the abstract plan is applicable. Based on this evaluation, several different indexing and retrieval mechanisms have been developed. In these retrieval procedures those abstract cases are preferred which have caused the most reduction in search during previous problem solving episodes. In particular, abstract cases which turn out to be useless for many concrete problems may even become completely removed from the case-base. The spectrum of developed retrieval approaches ranges from simple sequential search, via hierarchical clustering up to a sophisticated approach for balancing a hierarchy of abstract cases according to the statistical distribution of the cases within the problem space and their evaluated utility. More details on the generalization procedure can be found in (Bergmann, 1992a), while the evaluation and retrieval mechanisms are reported in (Bergmann & Wilke, 1994; Wilke, 1994). The whole multi-strategy system including the various interactions of the described components will be the topic of a forthcoming article, while first ideas can already be found in (Bergmann, 1992b, 1993). However, as the target of this paper we will concentrate on the core of PARIS, namely the approach to abstraction and refinement.

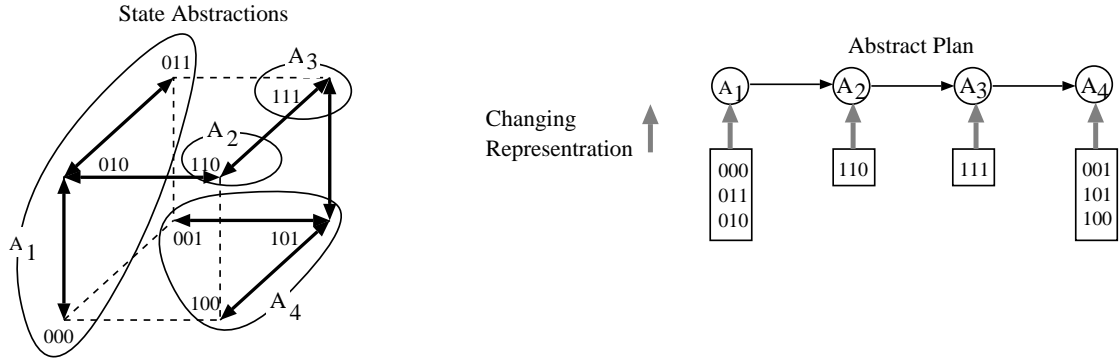

Figure 4: An example of case abstraction

## 3.3 Informal Description of the Abstraction Approach

We first give an informal description of the abstraction approach in PARIS, based on our small example shown in Figure 1 to enhance the understanding of the subsequent formal sections. Suppose that the solution to the problem $X \to X'$ is available as a concrete problem solving experience. The task is now to learn an abstract case which can be beneficially used to solve future problems such as $Y \to Y'$ and $Z \to Z'$. This learning task must be achieved within an abstraction approach which is stronger than dropping sentences. If we look at Figure 4, it becomes obvious that by changing the representation a single abstract case can be learned which is useful for all three concrete problems. The abstract plan shown indicates which concrete states have to be abstracted towards a single abstract state, such that *a single* abstract plan exists which is useful for all three problems.

### 3.3.1 Abstract Language and Generic Abstraction Theory

To achieve this kind of abstraction, our approach requires that the *abstract language* (states and operators), as well as a *generic abstraction theory* is provided by the user. For the example in Figure 4, the abstract language must contain the new abstract sentences $A_1, \ldots, A_4$ and the three abstract operators which allow the respective state transitions. These abstract operators, called $Oa_i$ ($i \in \{1, \ldots, 3\}$), can be defined as follows:

> *Operator: $Oa_i$*
> *Precondition: $A_i$*
> *Delete: $A_i$*
> *Add: $A_{i+1}$*

For each new abstract sentence, the user must provide a set of generic abstraction rules which describe how this sentence is defined in terms of the available sentences of the con-crete language. The generic abstraction theory defined by these rules specifies a set of admissible state abstractions. For our example, the generic abstraction theory must contain the following two rules to define the new abstract sentence $A_1$: $\neg E1 \wedge E2 \rightarrow A_1$ and $\neg E1 \wedge \neg E2 \wedge \neg E3 \rightarrow A_1$. In general, the definition of the generic abstraction theory does not require that all state abstractions are noted explicitly. Abstract states can be derived implicitly by the application of a combination of several rules from the generic abstraction theory.

Besides the kind of abstraction described above, the user may also want to specify a different type of abstraction which she/he also considers useful. For example, we can assume that abstraction by dropping the sentence $E1$ should also be realized. In this case, the abstract language must contain a copy of the two sentences which are not dropped, i.e., the sentences $E2$ and $E3$. Therefore, the user[5] may define two abstract sentences $A_5$ and $A_6$ by the following rules of the generic abstraction theory: $E2 \rightarrow A_5$ and $E3 \rightarrow A_6$. Of course, the respective abstract operators must also be specified.

Since the domain expert or knowledge engineer must provide the abstract language and the generic abstraction theory, she/he must already have one or more particular kinds of abstraction in mind. She/he must know what kind of details can be omitted when solving a problem in an abstract fashion. With our approach, the knowledge-engineer is given the power to express the kind of abstraction she/he considers useful.

### 3.3.2 Model of Case Abstraction

Based on the given abstract language and the generic abstraction theory, the abstraction of a planning case can be formally described by two abstraction mappings: a *state abstraction mapping* and a *sequence abstraction mapping*. These two mappings describe two dimensions for reducing the level of detail in a case. The state abstraction mapping reduces the level of detail of a state description while changing the representation language. For the case abstraction indicated in Figure 4, the state abstraction mapping must map the concrete states 000, 011 and 010 onto an abstract state described by the new sentence $A_1$, and simultaneously it must map all other concrete states occurring in the plan onto the respective abstract states described by the new sentences $A_2$, $A_3$, and $A_4$. The sequence abstraction mapping reduces the level of detail in the number of states which are considered at the abstract level by relating *some* of the concrete states from the concrete case to abstract states of the abstract case. While some of the concrete states can be skipped, each abstract state must result from a particular concrete state. For example, in Figure 4, the abstraction of the plan $000 \rightarrow 010 \rightarrow 110 \rightarrow 111 \rightarrow 101 \rightarrow 001$ requires a sequence abstraction mapping which relates the first abstract state described by $A_1$ to the first concrete state 000, the second abstract state described by $A_2$ to the third concrete state 110, and so forth. In this example, the second and the fifth concrete states are skipped.

### 3.3.3 Learning Abstract Planning Cases

The procedure for learning such abstract planning cases from a given concrete planning case is decomposed into four separate phases. For our simple example, these phases are shown in the transitions shown in plain line style result from the operators $Oa_i$, while the transitions shown in dashed line style result from the operators required for abstraction by dropping conditions.

---

5. Please note that for abstraction by dropping sentences, we can also consider an ALPINE-like algorithm which generates the required abstract language and the generic abstraction theory automatically.

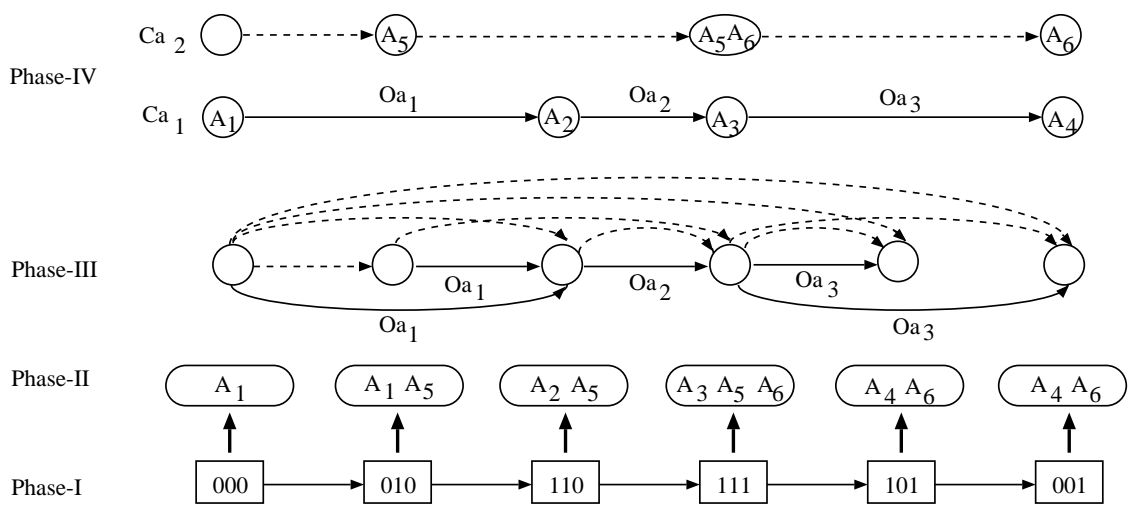

Figure 5: The four phases of case abstraction for the solution to the problem $X \rightarrow X'$

Figure 5. In phase-I, the states which result from the execution of the plan contained in the concrete case are determined. Therefore, each operator contained in the plan (starting from the first operator) is applied and the successor state is computed. This process starts at the initial state contained in the case and leads to a final state, which should be the goal state contained in the case. In phase-II, we derive all admissible abstractions for *each* concrete state computed in the first phase. For this purpose, the generic abstraction theory is used to determine all abstract sentences that can be derived from a respective concrete state by applying the rules of the generic abstraction theory. Figure 5 shows the abstract sentences that can be derived by the generic abstraction theory sketched above. For example, we can see that for the second concrete state an abstract description can be derived which contains two abstract sentences: the abstract sentence $A_1$ required to achieve the type of abstraction shown in Figure 4 and additionally the abstract sentences $A_5$ required for abstraction by dropping sentences. Please note that by this process, the representation language of states is changed from concrete to abstract. The next two phases deal with the abstract operators. As already stated, abstract operators are given in the abstract language provided by the user. However, we do not assume operator abstraction rules which associate an abstract operator to a single concrete operator or a sequence of concrete operators. The reason for this is that such operator abstraction rules are extremely hard to acquire and even harder to keep complete. In the next two phases of case abstraction, we search for transitions of abstract states based on the available abstract operators. In phase-III, an acyclic directed graph is constructed. An edge leads from an abstract state $i$ to a successor abstract state $j$ (not necessarily to the next abstract state), if the abstract operator is applicable in state $i$ and its application leads to the state $j$. The definition of the abstract operators are used in this process. The available abstract operators determine which transitions can be included in the graph. Figure 5 shows the resulting graph, provided that the abstract operators sketched in Section 3.3.1 are contained in the abstract language. In this graph

In phase-IV the graph is searched for consistent paths from the initial abstract state to the final abstract state. The paths must be consistent in the sense that in the resulting path (i.e., an abstract plan) every abstract operator is correctly applicable in the state that results from the previous operator. Moreover, the state abstraction which is required for this abstract plan must not change within the plan. In Figure 5 two paths of this kind are shown. The lower path represents the abstract planning case $Ca_1$ (abstract initial and final state together with the operator sequence) that results from the kind of abstraction shown in Figure 4. The upper path represents the abstract planning case $Ca_2$ that results from abstraction by dropping the sentence $E1$. This is the same abstract plan as shown in Figure 2 for the problem $X \to X'$. Together with the two plans, the abstract state descriptions that result from the operator application are shown. Please note that these state descriptions are always a subset of the description which are derived by the generic abstraction theory. For example, the description of the fourth abstract state derived in phase-II, contains the sentences $A_3, A_5, A_6$. This abstract state occurs in both abstract cases which are computed in phase-IV. In the case $Ca_2$, the respective state is described by the sentences $A_5$ and $A_6$ because these are the only sentences which result from the application of the operators starting at the abstract initial state. In the case $Ca_1$, the abstract state is described by the sentence $A_3$ because this sentence results from the application of the operator $Oa_2$. From this example we can see that the abstract operators have two functions. The first function is to select some of the concrete states that become abstracted. For example, in the abstract case $Ca_1$, the second concrete state is skipped, even if the first and the second concrete states can be abstracted to different abstract descriptions in phase-II. The reason for this is that there is no abstract operator that a) leads from the first abstracted state to the second abstracted state and which b) is also consistent with the other operators in the rest of the path. The second function of the abstract operators is to select some of the abstract sentences that are considered in the abstract planning case. For example, in the abstract case $Ca_1$, the sentences $A_1, \ldots, A_4$ are considered while the sentences $A_5$ and $A_6$ are left out. The reason for this is that the abstract operators $Oa_1, Oa_2, Oa_3$ which occur in the plan don't use $A_5$ and $A_6$ in their precondition and don't manipulate these sentences.

After phase-IV is finished, a set of abstract planning cases is available. These planning cases can then be stored in the case-base and used for further problem solving.

### 3.3.4 SELECTING AND REFINING ABSTRACT CASES

During problem solving, an abstract case must be selected from a case-base, and the abstract plan contained in this case must be refined to become a solution to the current problem. During case retrieval we must search for an abstract case which is *applicable*, i.e., it contains a problem description that is an abstraction of the current problem. For example, assume that the problem $Y \to Y'$ should be solved after the case $X \to X'$ was presented for learning. In this situation the case-base contains the two abstract cases $Ca_1$ and $Ca_2$ shown in phase-IV of Figure 5. The abstract case $Ca_1$ can be used for solving the new problem, because the initial state 000 of the new problem can be abstracted to $A_1$ by applying the generic abstraction theory. Similarly, the final state 100 can be abstracted to $A_4$. However, the abstract case $Ca_2$ is not applicable because the final abstract state cannot be abstracted to $A_6$. Consequently, the lower abstract case must be used. During plan refinement we can refine the abstract operators sequentially from left to right as shown in Figure 6. Thereby each abstract operator defines an abstract goal state, i.e., the state that results after the execution of the operator. For example, the abstract operator $Oa_1$ defines the abstract goal $A_2$. To refine an abstract operator, we search for a concrete operator sequence, starting from the current concrete state (i.e., the initial state for the first operator), until a concrete state is reached that can be abstracted to the desired goal state. If such a state is found it can be used as a starting state for the refinement of the next abstract operator. For the solution of the problem $Y \rightarrow Y'$, the refinement of the abstract operator $Oa_1$ can be achieved by a sequence of two concrete operators leading to the concrete state 110. This concrete state is then used as a starting state to refine the next abstract operator $Oa_2$. This refinement procedure finishes if the last abstract operator is refined in a way that the final concrete state is achieved. Please note that in this type of refinement the operators themselves are not used directly, instead the sequence of states which results from their execution are used. Alternatively, we could have also stored an abstract case as a sequence of abstracted states. From our experience, storing a sequence of operators requires less space than storing a sequence of states. This will become obvious when looking at the domain that will be introduced in Section 8. Besides this the abstract operators play an important role in the learning phase.

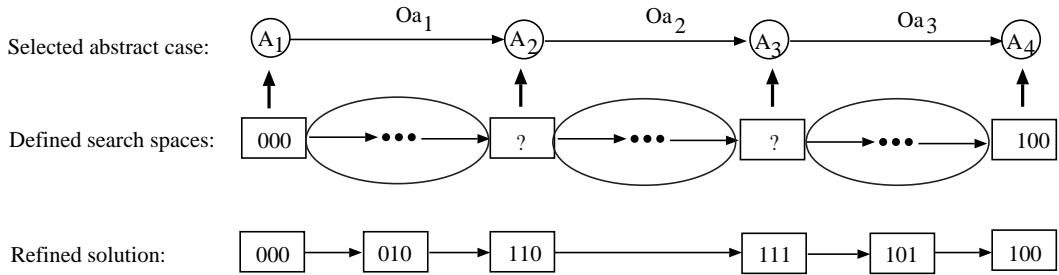

Figure 6: Refinement of an abstract case for the solution of the problem $Y \rightarrow Y'$

### 3.4 Relations to Skeletal Plans

A similar experience-based or case-based variant for finding an abstract solution can be found in an early paper by Friedland and Iwasaki (1985) in which the concept of *skeletal plans* is introduced. A skeletal plan is "*[...] a sequence of generalized steps, which, when instantiated by specific operations in a specific problem context will solve a given problem [p.161]. [...] Skeletal plans exist at many levels of generality. At the most general level, they are only a few basic plans, but these are used as 'fall-backs', when more specific, easier to refine plans cannot be found. [p. 164].*" Skeletal plans are solutions to planning problems at different levels of detail and are consequently *abstract plans*. During problem solving they are recalled from a library and refined towards a concrete solution. So this approach can be seen as an early idea for integrating abstraction and case-based reasoning. However, there are several differences between the skeletal plan approach and the PARIS approach. In the skeletal plan approach no model of the operators (neither concrete nor abstract) is used to describe the preconditions and effects of operators as is done in PARIS. There is no explicit notion of states and abstraction or refinement of states. Instead, the plan refinement is achieved by stepping down a hierarchy of operators, guided by heuristic rules for operator selection. In particular, no approach which supports the automatic acquisition of skeletal plans was provided. Unfortunately, the skeletal plan approach has not yet been investigated in as much detail as current work in the field of speedup-learning. There is neither a formal model of skeletal planning nor empirical evaluations.

In the rest of this paper we will introduce and investigate the PARIS approach more formally.

## 4. Basic Terminology

In this section we want to introduce the basic formal terminology used throughout the rest of this paper. Therefore we will define a formal representation for problem solving domains. We want to assume that problem solving in general can be viewed as transforming an *initial state* into a *final state* by using a sequence of *operators* (Newell & Simon, 1972). Following a STRIPS-oriented representation (Fikes & Nilsson, 1971), the domain of problem solving $\mathcal{D} = \langle \mathcal{L}, \mathcal{E}, \mathcal{O}, \mathcal{R} \rangle$ is described by a first-order language[6] $\mathcal{L}$, a set of essential atomic sentences $\mathcal{E}$ of $\mathcal{L}$ (Lifschitz, 1987), a set of operators $\mathcal{O}$ with related descriptions, and additionally, a set of rules (Horn clauses) $\mathcal{R}$ out of $\mathcal{L}$. The essential sentences (which must be atomic) are the only sentences that are used to describe a state. A state $s \in \mathcal{S}$ describes the dynamic part of a situation in a domain and consists of a finite subset of ground instances of essential sentences of $\mathcal{E}$. With the symbol $\mathcal{S}$, we denote the set of all possible states descriptions in a domain, which is defined as $\mathcal{S} = 2^{\mathcal{E}^*}$, with $\mathcal{E}^* = \{e\sigma | e \in \mathcal{E}$ and $\sigma$ is a substitution such that $e\sigma$ is ground$\}$. In addition, the Horn clauses $\mathcal{R}$ allow the representation of static properties which are true in all situations. These Horn clauses must not contain an essential sentence in the head of a clause.

An operator $o(x_1, \ldots, x_n) \in \mathcal{O}$ is described by a triple $\langle Pre_o, Add_o, Del_o \rangle$, where the precondition $Pre_o$ is a conjunction of atoms of $\mathcal{L}$, and the add-list $Add_o$ and the delete-list $Del_o$ are finite sets of (possibly instantiated) essential sentences of $\mathcal{E}$. Furthermore, the variables occuring in the operator descriptions must follow the following restrictions: $\{x_1, \ldots, x_n\} \supseteq Var(Pre_o) \supseteq Var(Del_o)$ and $\{x_1, \ldots, x_n\} \supseteq Var(Add_o)$.[7]

An *instantiated* operator is an expression of the form $o(t_1, \ldots, t_n)$, with $t_i$ being ground terms of $\mathcal{L}$. A term $t_i$ describes the instantiation of the variable $x_i$ in the operator description. For notational convenience we define the *instantiated precondition* as well as the *instantiated add-list* and *delete-list* for an instantiated operator as follows: $Pre_{o(t_1, \ldots, t_n)} := Pre_o\sigma$, $Add_{o(t_1, \ldots, t_n)} := \{a\,\sigma | a \in Add_o\}$, $Del_{o(t_1, \ldots, t_n)} := \{d\,\sigma | d \in Del_o\}$, with $\langle Pre_o, Add_o, Del_o \rangle$ is

---

6. The basic language is first order, but with the deductive rules given in Horn logic only a subset of the full first-order language is used.

7. These restrictions can however be relaxed such that $\{x_1, \ldots, x_n\} \supseteq Var(Pre_o)$ is not required. But the introduced restriction simplifies the subsequent presentation.

the description of the (uninstantiated) operator $o(x_1, \ldots, x_n)$, and $\sigma = \{x_1/t_1, \ldots, x_n/t_n\}$ is the corresponding instantiation.

An instantiated operator $o$ is *applicable* in a state $s$, if and only if $s \cup \mathcal{R} \vdash Pre_o$ holds.[8] An instantiated operator $o$ transforms a state $s_1$ into a state $s_2$ (we write: $s_1 \overset{o}{\longrightarrow} s_2$) if and only if $o$ is applicable in $s_1$ and $s_2 = (s_1 \setminus Del_o) \cup Add_o$. A *problem description* $p = \langle s_I, s_G \rangle$ consists of an initial state $s_I$ together with a final state $s_G$. The problem solving task is to find a sequence of instantiated operators (a *plan*) $\bar{o} = (o_1, \ldots, o_l)$ which transforms the initial state into the final state $(s_I \overset{o_1}{\longrightarrow} \cdots \overset{o_l}{\longrightarrow} s_G)$. A *case* $C = \langle p, \bar{o} \rangle$ is a problem description $p$ together with a plan $\bar{o}$ that solves $p$.

The introduced STRIPS-oriented formalism for defining a problem solving domain is similar in form and expressiveness to the representations typically used in general problem solving or planning. A state can be described by a finite set of ground atoms in which functions can also be used. Full Horn logic is available to describe static rules. The restriction to Horn clauses has the advantage of being powerful while allowing efficient proof construction by using the well known SLD-refutation procedures (Lloyd, 1984). Compared to the Prodigy Description Language (PDL) (Minton, 1988; Blythe et al., 1992) our language does not provide explicit quantification by a specific syntactic construct, but a similar expressiveness can be reached by the implicit quantification in Horn clauses. Moreover, our language does not provide any kind of type specification for constants or variables as in PDL but we think that this is not a major disadvantage. Besides these points our language is quite similar to PDL.

## 5. A Formal Model of Case Abstraction

In this section we present a new formal model of case abstraction which provides a theory for changing the representation language of a case from concrete to abstract. As already stated we assume that in addition to the concrete language the abstract language is supplied by a domain expert. Following the introduced formalism, we assume that the concrete level of problem solving is defined by a *concrete problem solving domain* $\mathcal{D}_c = \langle \mathcal{L}_c, \mathcal{E}_c, \mathcal{O}_c, \mathcal{R}_c \rangle$ and the abstract level of (case-based) problem solving is represented by an *abstract problem solving domain* $\mathcal{D}_a = \langle \mathcal{L}_a, \mathcal{E}_a, \mathcal{O}_a, \mathcal{R}_a \rangle$. For reasons of simplicity, we assume that both domains do not share the same symbols[9]. This condition can always be achieved by renaming symbols. In the remainder of this paper states and operators from the concrete domain are denoted by $s^c$ and $o^c$ respectively, while states and operators from the abstract domain are denoted by $s^a$ and $o^a$ respectively. The problem of case abstraction can now be described as transforming a case from the concrete domain $\mathcal{D}_c$ into a case in the abstract domain $\mathcal{D}_a$ (see Figure 7). This transformation will now be formally decomposed into two independent mappings: a *state abstraction mapping* $\alpha$, and a *sequence abstraction mapping* $\beta$ (Bergmann, 1992c). The state abstraction mapping transforms a selection of concrete state descriptions that occur in the solution to a problem into abstract state descriptions, while the sequence abstraction mapping specifies which of the concrete states are mapped and which are skipped.

---

8. In the following, we will simply omit the parameters of operators and instantiated operators in case they are unambiguous or not relevant.

9. Otherwise, a symbol (or a sentence) could become ambiguous which would be a problem when applying the generic abstraction theory. It would be unclear whether a generic abstraction rule refers to a concrete or an abstract sentence

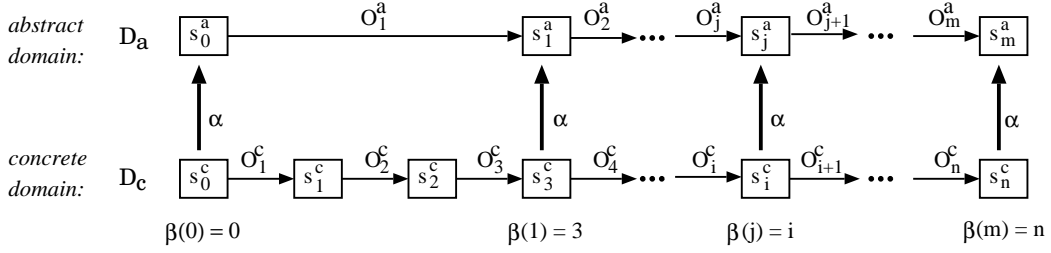

Figure 7: General idea of abstraction

## 5.1 State Abstraction

A state abstraction mapping translates states of the concrete world into the abstract world.

**Definition 1** (State Abstraction Mapping) *A state abstraction mapping $\alpha : \mathcal{S}_c \to \mathcal{S}_a$ is a mapping from $\mathcal{S}_c$, the set of all states in the concrete domain, to $\mathcal{S}_a$, the set of all states in the abstract domain. In particular, $\alpha$ must be an effective total function.*

This general definition of a state abstraction mapping does not impose any restrictions on the kind of abstraction besides the fact that the mapping must be a total many-to-one function. However, to restrict the set of all possible state abstractions to a set of abstractions which a user considers useful, we assume that additional domain knowledge about how an abstract state relates to a concrete state can be provided. This knowledge must be expressed in terms of a domain specific *generic abstraction theory $\mathcal{A}$* (Giordana, Roverso, & Saitta, 1991).

**Definition 2** (Generic Abstraction Theory) *A generic abstraction theory is a set of Horn clauses of the form $e_a \leftarrow a_1, \ldots, a_k$. In these rules $e_a$ is an abstract essential sentence, i.e., $e_a = E_a\sigma$ for $E_a \in \mathcal{E}_a$ and a substitution $\sigma$. The body of a generic abstraction rule consists of a set of sentences from the concrete or abstract language, i.e., $a_i$ are atoms out of $\mathcal{L}_c \cup \mathcal{L}_a$.*

Based on a generic abstraction theory, we can restrict the set of all possible state abstraction mappings to those which are deductively justified by the generic abstraction theory.

**Definition 3** (Deductively Justified State Abstraction Mapping) *A state abstraction mapping $\alpha$ is deductively justified by a generic abstraction theory $\mathcal{A}$, if the following conditions hold for all $s^c \in \mathcal{S}_c$:*

- *if $\phi \in \alpha(s^c)$ then $s^c \cup \mathcal{R}_c \cup \mathcal{A} \vdash \phi$ and*

- *if $\phi \in \alpha(s^c)$ then for all $\tilde{s}^c$ such that $\tilde{s}^c \cup \mathcal{R}_c \cup \mathcal{A} \vdash \phi$ holds, $\phi \in \alpha(\tilde{s}^c)$ is also fulfilled.*

In this definition the first condition assures that every abstract sentence reached by the mapping is justified by the abstraction theory. Additionally, the second requirement guarantees that if an abstract sentence is used to describe an abstraction of one state, it must also be used to describe the abstraction of all other states, if the abstract sentence can be derived by the generic abstraction theory. Please note that a deductively justified state abstraction mapping can be completely induced by a set $\alpha^* \subseteq \mathcal{E}_a^*$ with respect to a generic abstraction theory as follows: $\alpha(s^c) := \{\phi \in \alpha^* | s^c \cup \mathcal{R}_c \cup \mathcal{A} \vdash \phi\}$. Unless otherwise stated we always assume deductively justified state abstraction mappings. To summarize, the state abstraction mapping transforms a concrete state description into an abstract state description and thereby changes the representation of a state from concrete to abstract. Please note that deductively justified state abstraction mappings need not to be defined by the user. They will be determined automatically by the learning algorithm that will be presented in Section 6.

## 5.2 Sequence Abstraction

The solution to a problem consists of a sequence of operators and a corresponding sequence of states. To relate an abstract solution to a concrete solution, the relationship between the abstract states (or operators) and the concrete states (or operators) must be captured. Each abstract state must have a corresponding concrete state but not every concrete state must have an associated abstract state. This is due to the fact that abstraction is always a reduction in the level of detail (Michalski & Kodratoff, 1990), in this situation, a reduction in the number of states. For the selection of those concrete states that have a corresponding abstraction, the sequence abstraction mapping is defined as follows:

**Definition 4** (Sequence Abstraction Mapping) *A sequence abstraction mapping $\beta : \mathbb{N} \rightarrow \mathbb{N}$ relates an abstract state sequence $(s_0^a, \ldots, s_m^a)$ to a concrete state sequence $(s_0^c, \ldots, s_n^c)$ by mapping the indices $j \in \{1, \ldots, m\}$ of the abstract states $s_j^a$ into the indices $i \in \{1, \ldots, n\}$ of the concrete states $s_i^c$, such that the following properties hold:*

- *$\beta(0) = 0$ and $\beta(m) = n$: The initial state and the goal state of the abstract sequence must correspond to the initial and goal state of the respective concrete state sequence.*

- *$\beta(u) < \beta(v)$ if and only if $u < v$: The order of the states defined through the concrete state sequence must be maintained for the abstract state sequence.*

Note that the defined sequence abstraction mapping formally maps indices from the abstract domain into the concrete domain. As an abstraction mapping it should better map indices from the concrete domain to indices in the abstract domain, such as the inverse mapping $\beta^{-1}$ does. However, such a mapping is more inconvenient to handle formally since the range of definition of $\beta^{-1}$ must always be considered. Therefore we stick to the presented definition.

## 5.3 Case Abstraction

Based on the two abstraction functions introduced, our intuition of case abstraction is captured in the following definition.

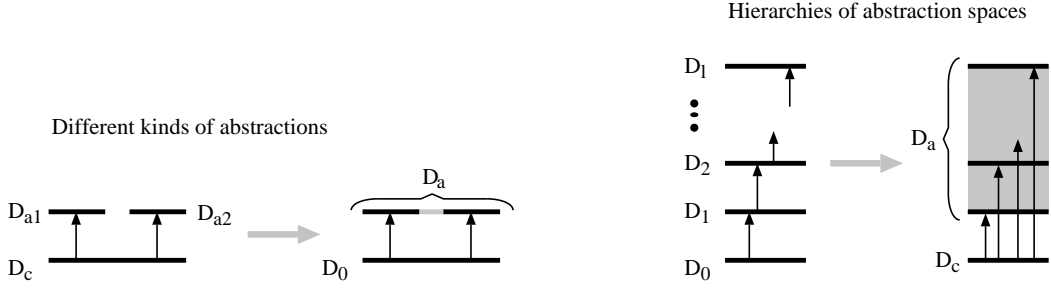

Figure 8: Different kinds of abstractions (a) and abstraction hierarchies (b)

**Definition 5** (Case Abstraction) *A case* $C_a = \langle\langle s_0^a, s_m^a\rangle, (o_1^a, \ldots, o_m^a)\rangle$ *is an abstraction of a case* $C_c = \langle\langle s_0^c, s_n^c\rangle, (o_1^c, \ldots, o_n^c)\rangle$ *with respect to the domain descriptions* $(\mathcal{D}_c, \mathcal{D}_a)$ *if* $s_{i-1}^c \xrightarrow{o_i^c} s_i^c$ *for all* $i \in \{1, \ldots, n\}$ *and* $s_{j-1}^a \xrightarrow{o_j^a} s_j^a$ *for all* $j \in \{1, \ldots, m\}$ *and if there exists a state abstraction mapping* $\alpha$ *and a sequence abstraction mapping* $\beta$, *such that:* $s_j^a = \alpha(s_{\beta(j)}^c)$ *holds for all* $j \in \{0, \ldots, m\}$.

This definition of case abstraction is demonstrated in Figure 7. The concrete space shows the sequence of $n$ operations together with the resulting state sequence. Selected states are mapped by $\alpha$ into states of the abstract space. The mapping $\beta$ maps the indices of the abstract states back to the corresponding concrete states.

## 5.4 Generality of the Case Abstraction Methodology

In the following, we briefly discuss the generality of the presented case abstraction methodology. We will see that hierarchies of abstraction spaces as well as different kinds of abstractions can be represented simultaneously using the presented methodology.

### 5.4.1 DIFFERENT KINDS OF ABSTRACTIONS

In general, there will be more than one possible abstraction of an object in the world. Abstraction can be performed in many different ways. An example of two different abstractions of the same case has already been shown in the example in Figure 5. In this example, two different abstractions (see the abstract cases $Ca_1$ and $Ca_2$) have been derived from the same concrete case. Our abstraction methodology is able cope with different abstractions in case they are specified by the user. Assume we are given one concrete domain $\mathcal{D}_c$ and two different abstract domains $\mathcal{D}_{a1}$ and $\mathcal{D}_{a2}$, each of which represents two different kinds of abstraction. Furthermore, assume that both abstract domains do not share the same symbols[10]. We can always define a single abstract domain $\mathcal{D}_a$ by joining the individual abstract domains which then includes both kinds of abstractions (see Figure 8 (a)). This property is formally captured in the following simple lemma.

---

10. If the abstract domains are not disjoint, symbols can be simply renamed to achieve this property.

**Lemma 6** (Joining different abstractions) *If a concrete domain $\mathcal{D}_c$ and two disjoint abstract domains $\mathcal{D}_{a1}$ and $\mathcal{D}_{a2}$ are given, then a joint abstract domain $\mathcal{D}_a = \mathcal{D}_{a1} \cup \mathcal{D}_{a2}$ can be defined as follows: Let $\mathcal{D}_{a1} = (\mathcal{L}_{a1}, \mathcal{E}_{a1}, \mathcal{O}_{a1}, \mathcal{R}_{a1})$ and let $\mathcal{D}_{a2} = (\mathcal{L}_{a2}, \mathcal{E}_{a2}, \mathcal{O}_{a2}, \mathcal{R}_{a2})$. Then $\mathcal{D}_a = \mathcal{D}_{a1} \cup \mathcal{D}_{a2} = (\mathcal{L}_{a1} \cup \mathcal{L}_{a2}, \mathcal{E}_{a1} \cup \mathcal{E}_{a2}, \mathcal{O}_{a1} \cup \mathcal{O}_{a2}, \mathcal{R}_{a1} \cup \mathcal{R}_{a2})$. The joint abstract domain $\mathcal{D}_a$ fulfills the following property: if $C_a$ is an abstraction of $C_c$ with respect to $(\mathcal{D}_c, \mathcal{D}_{a1})$ or with respect to $(\mathcal{D}_c, \mathcal{D}_{a2})$, then $C_a$ is also an abstraction of $C_c$ with respect to $(\mathcal{D}_c, \mathcal{D}_a)$.*

### 5.4.2 Hierarchy of Abstraction Spaces

Most work on hierarchical problem solving assume a multi-level hierarchy of abstraction spaces for problem solving (e.g., Sacerdoti, 1974; Knoblock, 1989). Even if the presented approach contains only two domain descriptions, a hierarchy of abstract domains can simply be mapped onto the presented two-level model as shown in Figure 8 (b). Assume that a hierarchy of disjoint domain descriptions $(\mathcal{D}_0, \dots, \mathcal{D}_l)$ is given. In particular, the domain $\mathcal{D}_{\nu+1}$ is assumed to be more abstract than the domain $\mathcal{D}_\nu$. In such a multi-level hierarchy of abstraction spaces, a case $C_\nu$ at the abstraction level $\mathcal{D}_\nu$ is an abstraction of a case $C_0$, if there exists a sequence of cases $(\mathcal{C}_1, \dots, \mathcal{C}_{\nu-1})$ such that $\mathcal{C}_i$ is out of the domain $\mathcal{D}_i$ and $\mathcal{C}_{i+1}$ is an abstraction of $\mathcal{C}_i$ with respect to $(\mathcal{D}_i, \mathcal{D}_{i+1})$ for all $i \in \{0, \dots, \nu-1\}$. Such a multi-level hierarchy of domain descriptions can always be reduced to a two-level description. The abstract domain of this two-level description contains the union of all the levels from the multi-level hierarchy. This property is formally captured in the following lemma.

**Lemma 7** (Multi-Level Hierarchy) *Let $(\mathcal{D}_0, \dots, \mathcal{D}_l)$ be an arbitrary multi-level hierarchy of domain descriptions. For the two-level description $(\mathcal{D}_c, \mathcal{D}_a)$ with $\mathcal{D}_a = \bigcup_{\nu=1}^{l} \mathcal{D}_\nu$ and $\mathcal{D}_c = \mathcal{D}_0$ holds that: if $C_a$ is an abstraction of $C_c$ with respect to $(\mathcal{D}_0, \dots, \mathcal{D}_l)$ then $C_a$ is also an abstraction of $C_c$ with respect to $(\mathcal{D}_c, \mathcal{D}_a)$.*

Since we have shown that different kinds of abstractions as well as hierarchies of abstraction spaces can be directly represented within our two-level case abstraction methodology, we can further restrict ourselves to exactly these two levels.

## 6. Computing Case Abstractions

We now present the Pabs algorithm (Bergmann, 1992c; Wilke, 1993) for automatically learning a set of abstract cases from a given concrete case. Thereby, we assume that a concrete domain $\mathcal{D}_c$ and an abstract domain $\mathcal{D}_a$ are given together with a generic abstraction theory $\mathcal{A}$. We use the functional notation $C_a \in \text{PABS}(\langle \mathcal{D}_c, \mathcal{D}_a, \mathcal{A} \rangle, C_c)$ to denote that $C_a$ is an element of the set of abstract cases returned by the Pabs algorithm.

The algorithm consists of the four separate phases introduced in Section 3. In the following we will present these phases in more detail.

In the first three phases, we require a procedure for determining whether a conjunctive formula is a consequence of a set of Horn clauses. For this purpose, we use a SLD-refutation procedure (Lloyd, 1984) which is given a set of Horn clauses (a logic program) $C$ together with conjunctive formula $G$ (a goal clause). The refutation procedure determines a set of answer substitutions $\Omega$ such that $C \vdash G\sigma$ holds for all $\sigma \in \Omega$. We write $\Omega = \text{SLD}(C, G)$. This SLD-refutation procedure performs some kind of backward-chaining and works as follows. It selects a literal from the goal clause $G$ (i.e., the left most literal) and searches for a Horn clause in the logic program $C$ that contains a literal in its head that unifies with the selected goal literal. The selected literal is removed from $G$ and the body (if not empty) of the applied clause is added at the beginning of the goal clause. Then the most general unifier of the goal literal and the head of the clause is applied to the whole new goal clause. The resulting goal clause is called *resolvent*. This process continues until the goal clause becomes empty or until no more resolvents can be built. In the former case, the goal has been proven and an answer substitution is computed by composing the substitutions used during resolution. Backtracking is used to look for possible other selections of applicable Horn rules to determine alternative answer substitutions. The set of all answer substitutions is returned as set $\Omega$. If the whole space of possible applications of the available Horn rules has been searched unsuccessfully, the goal clause is not a consequence of the logic program $C$ and the SLD-refutation procedure terminates without an answer substitution ($\Omega = \emptyset$). This must not be confused with the situation in which an empty substitution is returned ($\Omega = \{\emptyset\}$), if no variables occur in $G$. In phase-III of the PABS algorithm, we will also require the derivation trees in addition to the answer substitutions. Then we write $\Pi = \text{SLD}(C, G)$ and assume that $\Pi$ is a set of pairs $(\sigma, \tau)$, where $\sigma$ is an answer substitution and $\tau$ is a derivation of $C \vdash G\sigma$.

In order to assure the termination of the SLD-refutation procedure we have to require that the abstract domain and the generic abstraction theory is designed according to the following principles[11]:

- For each concrete state $s^c \in \mathcal{S}_c$ and each concrete operator $o^c \in \mathcal{O}_c$ where $o^c$ is described by $\langle Pre_{o^c}, Add_{o^c}, Del_{o^c} \rangle$, $\text{SLD}(s^c \cup \mathcal{R}_c, Pre_{o^c})$ must lead to a finite set of ground substitutions of all variables which occur in $Pre_{o^c}$.

- For each state abstract $s^a \in \mathcal{S}_a$ and each abstract operator $o^a \in \mathcal{O}_a$ where $o^a$ is described by $\langle Pre_{o^a}, Add_{o^a}, Del_{o^a} \rangle$, $\text{SLD}(s^a \cup \mathcal{R}_a, Pre_{o^a})$ must lead to a finite set of ground substitutions of all variables which occur in $Pre_{o^a}$.

- For each state $s^c \in \mathcal{S}_c$ and each abstract essential sentence $E \in \mathcal{E}_a$, $\text{SLD}(s^c \cup \mathcal{R}_c \cup \mathcal{A}, E)$ must lead to a finite set of ground substitutions of all variables which occur in $E$.

In the following the four phases of the PABS algorithm are explained in detail.

## 6.1 Phase-I: Computing the Concrete State Sequence

As input to the case abstraction algorithm, we assume a concrete case $C_c = \langle \langle s_I^c, s_G^c \rangle, (o_1^c, \ldots, o_n^c) \rangle$. Note that $(o_1^c, \ldots, o_n^c)$ is a totally ordered sequence of *instantiated* operators similar to the plans in PRODIGY (Minton, 1988; Minton et al., 1989; Veloso & Carbonell, 1993). In the first phase, the state sequence which results from the simulation of problem solution is computed as follows:

---

11. At first glance, this restrictions seem a bit hard to achieve but if we take a closer look at it we can see that this is the standard requirement for a (terminating) logic program (i.e., a PROLOG program).

**Algorithm 1** (Phase-I: Computing the concrete state sequence)

$s_0^c := s_I^c$
**for** $i := 1$ **to** $n$ **do**
    **if** $\mathrm{SLD}(s_{i-1}^c \cup \mathcal{R}_c, Pre_{o_i^c}) = \emptyset$ **then** *STOP "Failure: Operator not applicable"*
    $s_i^c := (s_{i-1}^c \setminus Del_{o_i^c}) \cup Add_{o_i^c}$
    **end**
**if** $s_G^c \not\subseteq s_n^c$ **then** *STOP "Failure: Goal state not reached"*

By this algorithm, the states $s_i^c$ are computed, such that $s_{i-1}^c \xrightarrow{o_i^c} s_i^c$ holds for all $i \in \{1, \ldots, n\}$. If a failure occurs the given plan is not valid, i.e., it does not solve the given problem.

## 6.2 Phase-II: Deriving Abstract Essential Sentences

Using the derived concrete state sequence as input, the following algorithm computes a sequence of abstract state descriptions $(s_i^a)$ by applying the generic abstraction theory separately to each concrete state.

**Algorithm 2** (Phase-II: State abstraction)

**for** $i := 0$ **to** $n$ **do**
    $s_i^a := \emptyset$
    **for each** $E \in \mathcal{E}_a$ **do**
        $\Omega := \mathrm{SLD}(s_i^c \cup \mathcal{R}_c \cup \mathcal{A}, E)$
        **for each** $\sigma \in \Omega$ **do**
            $s_i^a := s_i^a \cup \{E\,\sigma\}$
            **end**
        **end**
    **end**

Please note that we have claimed that the domain theories are designed in a way that $\Omega$ is finite and only contains ground substitution of all variables in $E$. Therefore, every description $s_i^a$ consists only of ground atoms and is consequently a valid abstract state description. Within the introduced model of case abstraction we have now computed a superset for the outcome of possible state abstraction mappings. Each deductively justified state abstraction mapping $\alpha$ is restricted by $\alpha(s_i^c) \subseteq s_i^a = \{e \in \mathcal{S}_a \,|\, s_i^c \cup \mathcal{R}_c \cup \mathcal{A} \vdash e\}$ for all $i \in \{1, \ldots, n\}$. Consequently, we have determined all abstract sentences that an abstract case might require.

## 6.3 Phase-III: Computing Possible Abstract State Transitions

In the next phase of the algorithm, we search for instantiated abstract operators which can transform an abstract state $\tilde{s}_i^a \subseteq s_i^a$ into a subsequent abstract state $\tilde{s}_j^a \subseteq s_j^a$ $(i < j)$. Therefore, the preconditions of the instantiated operator must at least be fulfilled in the state $\tilde{s}_i^a$ and consequently in also $s_i^a$. Furthermore, all added effects of the operator must be true in $\tilde{s}_j^a$ and consequently also in $s_j^a$.

**Algorithm 3** (Phase-III: Abstract state transitions)

$G := \emptyset$
**for** $i := 0$ **to** $n-1$ **do**
  **for** $j := i+1$ **to** $n$ **do**
    **for each** $o(x_1, \ldots, x_u) \in \mathcal{O}_a$ **do**
      **let** $\langle Pre_o, Del_o, Add_o \rangle$ *be the description of* $o(x_1, \ldots, x_u)$
      $\Pi := \text{SLD}(s_i^a \cup \mathcal{R}_a, Pre_o)$
      **for each** $\langle \sigma, \tau \rangle \in \Pi$ **do**
        **let** $Add_o' = \{a\sigma | a \in Add_o\}$
        (* Compute all possible instantiations  *)
        (* of added sentences which hold in $s_j^a$ *)
        $M := \{\emptyset\}$
        (* M is the set of possible substitutions *)
        (* initially the empty substitution. *)
        **for each** $a \in Add_o'$ **do**
          $M' := \emptyset$
          **for each** $\theta \in M$ **do**
            **for each** $e \in s_j^a$ **do**
              **if** *there is a substitution* $\rho$ *such that* $a\theta\rho = e$ **then** $M' := M' \cup \{\theta\rho\}$
              **end**
            **end**
          $M := M'$
        **end**
        (* Now, M contains the set of all possible substitutions *)
        (* such that all added sentences are contained in $s_j^a$ *)
        **for each** $\theta \in M$ **do**
          $G := G \cup \{\langle i, j, o(x_1, \ldots, x_u)\sigma\theta, \tau \rangle\}$
        **end**
      **end**
    **end**
  **end**
**end**

The set of all possible operator transitions are collected as directed edges of a graph which vertices represent the abstract states. In the algorithm, the set $G$ of edges of the acyclic directed graph is constructed. For each pair of states $(s_i^a, s_j^a)$ with $i < j$ it is checked whether there exists an operator $o(x_1, \ldots, x_u)\sigma$ which is applicable in $s_i^a$. For this purpose, the SLD-refutation procedure computes the set of all possible answer substitutions $\sigma$ such that the precondition of the operator is fulfilled in $s_i^a$. The derivation $\tau$ which belongs to each answer substitution is stored together with the operator in the graph since it is required for the next phase of case abstraction. This derivation is an "and-tree" where each inner-node reflects the resolution of a goal literal with the head of a clause and each leaf-node represents the resolution with a fact. Note that for proving the precondition of an abstract operator the inner nodes of the tree always refer to clauses of the Horn rule set $\mathcal{R}_a$, while the leave-nodes represent facts stated in $\mathcal{R}_a$ or essential sentences contained in $s_i^a$. Then each answer substitution $\sigma$ is applied to the add-list of the operator leading to a partially instantiated add-list $Add'_o$. Note that there can still be variables in $Add'_o$ because the operator may contain variables which are not contained in its precondition but may occur in the add-list. Therefore, the set $M$ of all possible substitutions $\theta$ is incrementally constructed such that $a\theta \in s_{aj}$ holds for all $a \in Add'_o$. The completely instantiated operator derived thereby is finally included as a directed edge (from $i$ to $j$) in the graph $G$.

By this algorithm it is guaranteed that each (instantiated) operator which leads from $s_i^a$ to $s_j^a$ is applicable in $s_i^a$ and that all essential sentences added by this operator are contained in $s_j^a$. Furthermore, if the applied SLD-refutation procedure is complete (it always finds all answer substitutions), then every instantiated operator which is applicable in $s_i^a$ such that all essential sentences added by this operator are contained in $s_j^a$ is also contained in the graph. From this follows immediately that if $\alpha(s_{\beta(i-1)}^c) \xrightarrow{o_i^a} \alpha(s_{\beta(i)}^c)$ holds for an arbitrary deductively justified state abstraction mapping $\alpha$ and a sequence abstraction mapping $\beta$, then $\langle \beta(i-1), \beta(i), o_i^a, \tau \rangle \in G$ also holds.

## 6.4 Phase-IV: Determining Sound Paths

Based on the state abstractions $s_i^a$ derived in phase-II and on the graph $G$ computed in the previous phase, phase-IV selects a set of sound paths from the initial abstract state to the final abstract state. A set of significant abstract sentences $\alpha^*$ and a sequence abstraction mapping $\beta$ are also determined during the construction of each path.

**Algorithm 4** (Phase-IV: Searching sound paths)[12]

$\quad Paths := \{\langle(), \emptyset, (\beta(0) = 0)\rangle\}$
$\quad$**while** *there exists* $\langle(o_1^a, \ldots, o_k^a), \alpha^*, \beta\rangle \in Paths$ *with* $\beta(k) < n$ **do**
$\quad\quad Paths := Paths \setminus \langle(o_1^a, \ldots, o_k^a), \alpha^*, \beta\rangle$
$\quad\quad$**for each** $\langle i, j, o^a, \tau \rangle \in G$ *with* $i = \beta(k)$ **do**
$\quad\quad\quad$**let** $\tau_\mathcal{E}$ *be the set of essential sentences contained in the derivation* $\tau$
$\quad\quad\quad$**let** $\alpha' = \tau_\mathcal{E} \cup Add_{o^a} \cup \alpha^*$
$\quad\quad\quad$**if for all** $\nu \in \{1, \ldots, k\}$ *holds:*
$\quad\quad\quad\quad (s_{\beta(\nu-1)}^a \cap \alpha') \xrightarrow{o_\nu^a} (s_{\beta(\nu)}^a \cap \alpha')$ **and**
$\quad\quad\quad\quad (s_{\beta(k)}^a \cap \alpha') \xrightarrow{o^a} (s_j^a \cap \alpha')$ **then**
$\quad\quad\quad\quad\quad Paths := Paths \cup \{\langle(o_1^a, \ldots, o_k^a, o^a), \alpha', \beta \cup \{\beta(k+1) = j\}\rangle\}$
$\quad\quad\quad$**end**
$\quad\quad$**end**
$\quad Cases_{Abs} := \emptyset$
$\quad$**for each** $\langle(o_1^a, \ldots, o_k^a), \alpha^*, \beta\rangle \in Paths$ *with* $\beta(k) = n$ **do**
$\quad\quad Cases_{Abs} := Cases_{Abs} \cup \{\langle\langle s_0^a \cap \alpha^*, s_n^a \cap \alpha^*\rangle, (o_1^a, \ldots, o_k^a)\rangle\}$
$\quad$**end**
$\quad$**return** $Cases_{Abs}$

While the construction of the sequence abstraction mapping is obvious, the set $\alpha^*$ represents the image of a state abstraction mapping $\alpha$ and thereby determines the set of sentences that have to be reached in order to assure the applicability of the constructed operator sequence. Note that from $\alpha^*$ the state abstraction mapping $\alpha$ can be directly determined as follows: $\alpha(s_i^c) = \{e \in \alpha^* | s_i^c \cup \mathcal{R}_c \cup \mathcal{A} \vdash e\}$. The idea of the algorithm is to start with an empty path. A path is extended by an operator from $G$ in each iteration of the algorithm until the path leads to the final state with the index $n$. New essential sentences $\alpha'$ may occur in the proof of the precondition or as added effects of this new operator. The path constructed so far must still be consistent according to the extension of the state description and, in addition, the new operator must transform the sentences of $\alpha^*$ correctly.

---

12. Please note that $\langle(o_1^a, \ldots, o_k^a), \alpha^*, \beta\rangle$ matches $\{\langle(), \emptyset, (\beta(0) = 0)\rangle\}$ with $k = 0$. The operator $\setminus$ denotes set difference.

As a result, phase-IV returns all cases that are abstractions of the given concrete input case with respect to concrete and abstract domain definitions and the generic abstraction theory. Depending on the domain theory, more than a single abstract case can be learned from a single concrete case as already shown in Figure 5.

### 6.5 Correctness and Completeness of the PABS Algorithm

Finally, we want to state again the strong connection between the formal model of case abstraction and the presented algorithm. The algorithm terminates if the domain descriptions and the generic abstraction theory are formulated as required in the beginning of this section, so that the SLD-resolution procedure always terminates. The algorithm is correct, that is every abstract case computed by the PABS algorithm is a case abstraction according to the introduced model. If the SLD-refutation procedure applied in PABS is complete every case which is an abstraction according to Definition 5 is returned by PABS. This property is captured in the following theorem.

**Theorem 8** (Correctness and completeness of the PABS algorithm) *If a complete SLD-refutation procedure is used in the* PABS *algorithm, then Case $C_a$ is an abstraction of case $C_c$ with respect to $(\mathcal{D}_c, \mathcal{D}_a)$ and the generic theory $\mathcal{A}$, if and only if $C_a \in \mathrm{PABS}(\langle \mathcal{D}_c, \mathcal{D}_a, \mathcal{A} \rangle, C_c)$.*

### 6.6 Complexity of the Algorithm

The complexity of the algorithm is mainly determined by the phases III and IV. The worst case complexity of phase-III is $O(n^2 \cdot C_1 \cdot C_2)$ where $n$ is the length of the concrete plan and $C_1$ and $C_2$ are dependent on the domain theories as follows: $C_1 = |\mathcal{O}_a| \cdot |\Omega|$ and $C_2 = |Add_{\mathcal{O}_a}| \cdot (|\mathcal{E}_a| \cdot |\Omega|)^{|Add_{\mathcal{O}_a}|}$. Thereby, $|\mathcal{O}_a|$ represents the number of abstract operators, $|\Omega|$ is the maximum number of substitutions found by the SLD-refutation procedure, $|Add_{\mathcal{O}_a}|$ is the maximum number of added sentences in an abstract operator, and $|\mathcal{E}_a|$ is the number of abstract essential sentences. The complexity of phase-IV can be determined as $O(n \cdot 2^{(n-1)} \cdot C_1)$. If we assume constant domain theories the overall complexity of the PABS algorithm can be summarized as $O(n \cdot 2^{(n-1)})$. The exponential factor comes from possibly exponential number of paths in a directed acyclic graph with $n$ nodes if every state is connected to every successor state. Whether a graph of this kind appears is very much dependent on the abstract domain theory, because it determines which transitions of abstract states are possible. This exponential nature does not lead to a time complexity problem in the domains we have used. Additionally, we want to make clear that this computational effort must be spent during learning and not during problem solving. If the time required for learning is very long, the learning phase can be executed off-line.

The space complexity of the algorithm is mainly determined by phase-III because all derivations of the proofs of the abstract operators' preconditions must be stored. This can sum up to $n^2 \cdot C_1 \cdot C_2$ derivations in the worst case. This did not turn out to be a problem in the domains we used because each derivation was very short (in most cases not more 3 inferences with static Horn rules). The reason for this is that the derivations relate to abstract operators which very likely contain less preconditions than the concrete operators.

## 7. Refinement of Abstract Cases

In the previous section we have described how abstract cases can be automatically learned from concrete cases. Now we assume a case-base which contains a set of abstract cases. We want to show how these abstract cases can be used to solve problems at the concrete level. Furthermore, we discuss the impact of the specific form of the abstract problem solving domain on the improvement in problem solving that can be achieved.

### 7.1 Applicability and Refinability of Abstract Cases

For a given abstract case and a concrete problem description, the question arises in which situations the abstract case can be *refined* to solve the concrete problem. For this kind of refinability an a-posterior definition can be easily given as follows.

**Definition 9** (Refinability of an abstract case) *An abstract case $C_a$ can be refined to solve a concrete problem $p$ if there exists a solution $\bar{o}_c$ to $p$, such that $C_a$ is an abstraction of $\langle p, \bar{o}_c \rangle$.*

Obviously, the refinability property is undecidable in general since otherwise planning itself would be decidable. However, we can define the *applicability* of an abstract case as a decidable necessary property for refinability as follows.

**Definition 10** (Applicability of an abstract case) *An abstract case $C_a = \langle \langle s_0^a, s_m^a \rangle, \langle o_1^a, \ldots, o_m^a \rangle \rangle$ can be applied to solve a concrete problem $p = \langle s_I^c, s_G^c \rangle$ if there exists a state abstraction mapping $\alpha$ such that $s_i^a \in \mathrm{Im}(\alpha)$ for all $i \in \{0, \ldots, m\}$ and $\alpha(s_I^c) = s_0^a$ and $\alpha(s_G^c) = s_m^a$. Thereby, $\mathrm{Im}(\alpha)$ denotes the image of the state abstraction mapping $\alpha$, i.e., all abstract states that can be reached.*

For an applicable abstract case, it is at least guaranteed that the concrete initial and goal states map to the abstract ones and that concrete intermediate states exists that can be abstracted as required by the abstract case.

Even if applicability is a necessary precondition for refinability it does not formally guarantee refinability, since the downward solution property (Tenenberg, 1988), which states that every abstract solution can be refined, is a too strong requirement to hold in general for our abstraction methodology. However, it is indeed guaranteed that each abstract case contained in the case-base is already an abstraction of one or more previous concrete cases due to the correctness of the PABS algorithm used for learning. If one of the problems contained in these concrete cases has to be solved again it is guaranteed that the learned abstract case can be refined to solve the problem. Consequently, each abstract case in the case-base can at least be refined to solve one problem that has occurred in the past.

Abstract solutions which are useless because they can never be refined to solve any concrete problem will never be in the case-base and are consequently never tried in solving a problem. Therefore, we expect that each abstract case from the case-base has a *high chance* of being also refinable for new similar problems for which it can be applied.

## 7.2 Selecting an Applicable Abstract Case

To decide whether an abstract case can be applied to solve a concrete problem $P$, we have to determine a suitable state abstraction mapping. Because we assume deductively justified state abstraction mappings, the required state abstraction mapping $\alpha$ can always be induced by the set $\alpha^* = \bigcup_{i=0}^{m} s_i^a$ as shown in Section 5.1. Consequently, $C_a$ is applicable to the problem $p = \langle s_I^c, s_G^c \rangle$ if and only if $s_0^a = \{\Phi \in \alpha^* \mid s_I^c \cup \mathcal{R}_c \cup \mathcal{A} \vdash \Phi\}$ and $s_m^a = \{\Phi \in \alpha^* \mid s_G^c \cup \mathcal{R}_c \cup \mathcal{A} \vdash \Phi\}$. Since every abstract case we use for solving a new problem has been learned from another concrete case, it is known that for each abstract state $s_i^a$ there must be at least one concrete state (from that previous concrete state) that can be abstracted via $\alpha$ to $s_i^a$. Consequently, $s_i^a \in \operatorname{Im}(\alpha)$ holds. Together with the introduced restrictions on the definition of $\mathcal{A}$ and $\mathcal{R}_c$ with respect to a complete SLD-refutation procedure (see Section 6), the applicability of an abstract case is decidable. Algorithm 5 describes the selection of an applicable abstract case for a problem $p = \langle s_I^c, s_G^c \rangle$ in more detail.

**Algorithm 5** (Selection of an applicable abstract case)

> $s_I^a := s_G^a := \emptyset$
> **for each** $E \in \mathcal{E}_a$ **do**
> > $\Omega := \operatorname{SLD}(s_I^c \cup \mathcal{R}_c \cup \mathcal{A}, E)$
> > $s_I^a := s_I^a \cup \bigcup_{\sigma \in \Omega} E\sigma$
> **for each** $E \in \mathcal{E}_a$ **do**
> > $\Omega := \operatorname{SLD}(s_G^c \cup \mathcal{R}_c \cup \mathcal{A}, E)$
> > $s_G^a := s_G^a \cup \bigcup_{\sigma \in \Omega} E\sigma$
> **repeat**
> > **repeat**
> > > *Select a new case $C_a = \langle \langle s_0^a, s_m^a \rangle, (o_1^a, \ldots, o_m^a) \rangle$ from the case base*
> > > *with $s_0^a \subseteq s_I^a$ and $s_m^a \subseteq s_G^a$*
> > > **if** *no more cases available* **then**
> > > > $refine_{DFID}(s_I^c, (), \emptyset, s_G^c)$
> > > > **return** *the result of $refine_{DFID}$*
> > > **for** $i := 1$ **to** $m - 1$ **do** $s_i^a := (s_{i-1}^a \setminus Del_{o_i^a}) \cup Add_{o_i^a}$
> > > $\alpha^* := \bigcup_{i=0}^{m} s_i^a$
> > **until** $(s_I^a \cap \alpha^*) = s_0^a$ *and* $(s_G^a \cap \alpha^*) = s_m^a$
> > $refine_{DFID}(s_I^c, (s_1^a, \ldots, s_{m-1}^a), \alpha^*, s_G^c)$
> **until** $refine_{DFID}$ *returns success(p)*
> **return** *success(p)*

At first, the initial and final concrete states of the problem are abstracted using the generic abstraction theory. Thereby, an abstract problem description $\langle s_I^a, s_G^a \rangle$ is determined. Then, in a pre-selection step, an abstract case is chosen form the case base. All of the abstract sentences contained in the initial and final abstract state of this case must be contained in the abstracted problem description $\langle s_I^a, s_G^a \rangle$. This condition, however, does not guarantee that the selected case is applicable with respect to Definition 10. The set $\alpha^*$ of abstract sentences inducing the respective state abstraction mapping is computed and the applicability condition is checked to test whether the selected case is applicable. If the selected case is not applicable, a new case must be retrieved. If an applicable abstract case has been determined the refinement algorithm $refine_{DFID}$ (see following section) is executed. This algorithm uses the sequence of intermediate abstract states $(s_1^a, \ldots, s_{m-1}^a)$, previously determined from the abstract plan of the case, to guide the search at the concrete level. The operators contained in the abstract plan are not used anymore. The refinement procedure returns $success(p)$, if the refinement succeeds with the solution plan $p$. If the refinement fails (the procedure returns $failure$), another case is selected. If no more cases are available the problem is solved by pure search without any guidance by an abstract plan.

## 7.3 Refining an Abstract Plan

The refinement of a selected abstract case starts with the concrete initial state from the problem statement. The search proceeds until a sequence of concrete operations is found which leads to a concrete state $s^c$, such that $s_1^a = \{\phi \in \alpha^* \mid s^c \cup \mathcal{R}_c \cup \mathcal{A} \vdash \phi\}$ holds. The applicability condition of the abstract case guarantees that such a state exists ($s_i^a \in \text{Im}(\alpha)$) but it is not guaranteed that the required concrete operator sequence exists too. Therefore, this search task may fail which causes the whole refinement process to fail also. If the first abstract operator can be refined successfully a new concrete state is found. This state can then be taken as a starting state to refine the next abstract operator in the same manner. If this refinement fails we can backtrack to the refinement of the previous operator and try to find an alternative refinement. If the whole refinement process reaches the final abstract operator it must directly search for an operator sequence which leads to the concrete goal state $s_G^c$. If this concrete goal state has been reached the concatenation of concrete partial solutions leads to a complete solution to original problem.

This refinement demands for a search procedure which allows an abstract goal specification. All kinds of forward-directed search such as depth-first iterative-deepening (Korf, 1985b) or best-first search (Korf, 1993) procedures can be used for this purpose because states are explicitly constructed during search. These states can then be tested to see if they can be abstracted towards the desired goal. In PARIS we use depth-first iterative-deepening search described by Algorithm 6. This algorithm consists of two recursive procedures. The top-level procedure $refine_{DFID}$ receives the concrete initial state $s_I^c$, the concrete final state $s_G^c$, the sequence of intermediate abstract states $S^a = (s_1^a, \ldots, s_k^a)$ derived from the abstract case, as well as the set $\alpha^*$ which induces the state abstraction mapping. This procedure increments the maximum depth for the depth-first search procedure $search_{bounded}$ up to the maximum $Deep_{Max}$. The procedure $search_{bounded}$ performs the actual search. The goal for this search is either an abstract state, i.e., the first abstract state in $S^a$, or the concrete goal state $s_G^c$ if all abstract state have already been visited. The procedure performs a depth-first search by applying the available concrete operators and recursively calling the search procedure with the concrete state $s_{new}^c$ which results from the operator application.

If an abstract goal state has been reached it is removed from the list $S^a$ and the refinement continues with the next abstract state which is then again the first one in the list.

**Algorithm 6** (Refinement by depth-first iterative-deepening (DFID) search)

> **procedure** $refine_{DFID}(s_I^c, S^a, \alpha^*, s_G^c)$
> $Deep := 0$
> **repeat**
> > $search_{bounded}(s_I^c, S^a, \alpha^*, s_G^c, Deep)$
> > **if** $search_{bounded}$ *returns success(p)* **then return** *success(p)*
> > $Deep := Deep + 1$ (* Search unsuccessful: Increment search deepness *)
> **until** $Deep = Deep_{Max}$
> **return** *failure*
>
> **procedure** $search_{bounded}(s_I^c, S^a, \alpha^*, s_G^c, Deep)$
> **if** $S^a = ()$ (* No more abstract goals: Test the concrete final goal *)
> > **and** $s_I^c = s_G^c$ **then return** *success(())*
> **if** $S^a = (s_1^a, \ldots, s_k^a)$ (* At least one abstract goal *)
> > **and** for all $e \in s_1^a$ *holds:* $\mathrm{SLD}(s_I^c \cup \mathcal{R}_c \cup \mathcal{A}, e) \neq \emptyset$
> > **and** for all $e \in \alpha^* \setminus s_1^a$ *holds:* $\mathrm{SLD}(s_I^c \cup \mathcal{R}_c \cup \mathcal{A}, e) = \emptyset$
> > **then** (* Abstract state reached: Refine next abstract operator *)
> > > $refine_{DFID}(s_I^c, (s_2^a, \ldots, s_k^a), \alpha^*, s_G^c)$
> > > **if** $refine_{DFID}$ *returns success(p)* **then return** *success(p)*
> **if** $Deep = 0$ **then return** *failure* (* Maximum depth reached *)
> (* Apply operators: Create successor states *)
> **for all** $o^c \in \mathcal{O}_c$ **do**
> > $\Omega = \mathrm{SLD}(s_I^c \cup \mathcal{R}_c, Pre_{o^c})$ (* $\Omega$ is the set of all possible operator instantiations *)
> > **for each** $\sigma \in \Omega$ **do**
> > > $s_{new}^c := (s_I^c \setminus (Del_{o^c}\sigma)) \cup (Add_{o^c}\sigma)$ (* Create successor state *)
> > > $search_{bounded}(s_{new}^c, S^a, \alpha^*, s_G^c, Deep - 1)$ (* Continue search with new state *)
> > > **if** $search_{bounded}$ *returns success(p)* **then return** $success((o^c\sigma) \circ p))$
> **return** *failure*

Please note that this kind of refinement is different from the standard notion of refinement in hierarchical problem solving (Knoblock et al., 1991b). This is because there is no strong correspondence between an abstract operator and a possible concrete operator. Moreover, the justification structure of a refined abstract plan is completely different from the justification structure of the abstract plan itself because of the completely independent definition of abstract and concrete operators. Even if this is a disadvantage compared to the usual refinement procedure used in hierarchical problem solving, the main computational advantage of abstraction caused by the decomposition of the original problem into smaller subproblems is maintained.

## 7.4 Alternative Search Procedures for Refinement

Besides the forward-directed search procedure currently used in PARIS backward-directed search as used in means-end analysis (Fikes & Nilsson, 1971) or in nonlinear partial-ordered planning (McAllester & Rosenblitt, 1991) can also be applied for refinement under certain circumstances. Therefore, we would either require a *state concretion function* or we have to turn the rules of the generic abstraction theory $\mathcal{A}$ into *virtual concrete operators*.

A state concretion function must be able to determine a single state or a finite set of concrete states from a given abstract state together with the concrete problem description. Thereby, the concrete problem description may help to reduce the number of possible concrete states. The derived state concretions can then be used as concrete goal states from which a backward directed search may start.

Alternatively, we can turn the process of state concretion directly into the search procedure by representing each rule in the generic abstraction theory as a virtual abstract operator. The precondition of a rule in the generic abstraction theory becomes the precondition of the virtual operator and the conclusion of the rule becomes a positive effect of this operator. When using the virtual concrete operators together with the operators of the concrete domain, a backward-directed planner can use the abstract state directly as a goal for search. The part of the plan in the resulting solution which only consists of concrete operators (and not of virtual operators) can be taken as a refinement of the abstract operator.

## 7.5 Criteria for Developing an Abstract Problem Solving Domain

The abstract problem solving domain and the generic abstraction theory used have an important impact on the improvement in problem solving that can be achieved. Therefore, it is desirable to have a set of criteria which state how a "good" abstract domain definition should look. Strong criteria allowing quantitative predictions of the resulting speedups can hardly be developed. For other hierarchical planners such criteria don't exist either. However, we can give a set of factors which determine the success of our approach. The overall problem solving time is influenced mainly by the following four factors: *independent refinability of abstract operators*, *goal distance of abstract operators*, *concrete scope of applicability of abstract operators*, and the *complexity of the generic abstraction theory*.

### 7.5.1 Independent Refinability of Abstract Operators

Following Korf's analysis of hierarchical problem solving (Korf, 1987) introduced in Section 2, our plan refinement approach reduces the overall search space from $b^n$ to $\sum_{i=1}^{m} b^{(\beta(i)-\beta(i-1))}$. Thereby, $b$ is the average branching factor, $n$ is the length of the concrete solution, and $\beta$ is the sequence abstraction mapping used in the abstraction of the concrete case to the abstract case. As already mentioned, we cannot guarantee that an abstract plan which is applicable to a problem can really be refined. Furthermore, Korf's analysis assumes that no backtracking between the refinement of the individual abstract operators is required which cannot be guaranteed. Some of the computational advantage of abstraction is lost in either of these two cases.

However, if the abstract operators occurring in the abstract problem solving domain fulfill the strong requirement of *independent refinability*, then it is guaranteed that every applicable abstract case can be refined without any backtracking. An abstract operator $o^a$ is independently refinable if for *each* $s^c$, $\tilde{s}^c \in \mathcal{S}_c$ and *every* state abstraction mapping $\alpha$ if $\alpha(s^c) \xrightarrow{o^a} \alpha(\bar{s}^c)$ holds, then there exists a sequence of concrete operators $(o_1^c, \ldots, o_k^c)$ such that $s^c \xrightarrow{o_1^c} \ldots \xrightarrow{o_k^c} \bar{s}^c$ holds.

The problem with this requirement is that it seems much to hard to develop an abstract problem solving domain in which all operators fulfill this requirement. Although we cannot expect that all operators in the abstract problem solving domain are independently refinable, a knowledge engineer developing an abstract domain should still try to define abstract operators which can be independently refined in *most* situations, i.e., for *most* $s^c, \bar{s}^c \in \mathcal{S}_c$ and *most* state abstraction mapping $\alpha$ an applicable abstract operator can be refined to a concrete operator sequence. Although this notion of mostly independent refinability is not formal we feel that it is practically useful when developing an abstract domain definition. The more abstract operators that can be refined independently in many situations, the higher is the chance that an abstract plan composed of these operators is also refinable.

### 7.5.2 GOAL DISTANCE OF ABSTRACT OPERATORS

The goal distance (cf. subgoal distance, Korf, 1987) is the maximum length of the sequence of concrete operators required to refine a particular abstract operator. The longer the goal distance the larger is the search space required to refine the abstract operator. In particular, the complexity of the search required to refine a complete abstract plan is determined by the largest goal distance of the abstract operators that occur in the abstract plan. Hence there is a good reason to keep the goal distance short. However, the goal distance negatively interacts with the next factor, namely the concrete scope of applicability of abstract operators.

### 7.5.3 CONCRETE SCOPE OF APPLICABILITY OF ABSTRACT OPERATORS

The concrete scope of applicability of an abstract operator specifies how many concrete states can be abstracted to an abstract state in which the abstract operator is applicable, and how many concrete states can be abstracted to an abstract state that can be reached by an abstract operator. This scope is determined by the definition of the abstract operator and by the generic abstraction theory which is responsible for specifying admissible state abstractions. The concrete scope of applicability of the abstract operators determines the applicability of the abstract plans that can be learned. An abstract plan which is only applicable to a few concrete problems is only of limited use in domains in which the problems to be solved vary very much. Hence, the concrete scope of applicability of abstract operators should be as large as possible. Unfortunately, according to our experience, abstract operators which have a large scope usually also have a larger goal distance and operators with a short goal distance don't have a large scope of applicability. Therefore, a compromise between these two contradicting issues must be found.

### 7.5.4 COMPLEXITY OF THE GENERIC ABSTRACTION THEORY

The fourth factor which influences the problem solving time is the complexity of the generic abstraction theory. This theory must be applied each time a new concrete state is created during concrete level search. The more complex the generic abstraction theory, the more time is required to compute state abstractions. Hence, the generic abstraction theory should not require complicated inferences and should avoid backtracking within the SLD-refutation procedure.

Although these four factors don't allow a precise prediction of the expected problem solving behavior of the resulting system, they provide a focus on what to consider when designing an abstract problem solving domain and related generic abstraction theory.

## 8. An Example Domain: Process Planning in Mechanical Engineering

The Paris approach has been successfully tested with toy-domains such as the familiar towers of Hanoi (Simon, 1975). For these domains, hierarchical problem solvers which use a dropping sentence approach have also proven very useful (Knoblock, 1994).

This section presents a new example domain we have selected from the field of process planning in mechanical engineering and which really requires a stronger abstraction approach.[13] We have selected the goal of generating a process plan for the production of a rotary-symmetric workpiece on a lathe. The problem description, which may be derived from a CAD-drawing, contains the complete specification (especially the geometry) of the desired workpiece (goal state) together with a specification of the piece of raw material (called mold) it has to be produced from (initial state).

The left side of Figure 9 shows an example of a rotary-symmetric workpiece which has to be manufactured out of a cylindrical mold.[14] Rotary parts are manufactured by putting the mold into the fixture (chuck) of a lathe. The chucking fixture, together with the attached mold, is then rotated with the longitudinal axis of the mold as rotation center. As the mold is rotated a cutting tool moves along some contour and thereby removes certain parts of the mold until the desired goal workpiece is produced. Within this process it is very hard to determine the sequence in which the specific parts of the workpiece have to be removed and the cutting tools to be used. When a workpiece is chucked a certain area of the workpiece is covered by the chucking tool and cannot be processed by a cutting tool. Moreover, a workpiece can only be chucked if the area which is used for chucking is plain. Otherwise the fixation would not be sufficiently stable. Hence, many workpieces are usually processed by first chucking the workpiece on one side and processing the accessible area. Then the workpiece is chucked at the opposite side and the area that was previously covered can be processed. Processing the example workpiece shown in Figure 9 requires that the workpiece is first chucked at the left side while the right side is processed. Then the processed right side can be used to chuck the workpiece because the area is plain and allows stable fixing. Hence, the left side of the workpiece including the small groove can be processed. Now we explain the representation of this domain in more detail. The complete definition of the domain can be found in Online Appendix 1. Several simplifications of the real domain were required in order to obtain a domain definition that could be efficiently handled in a large set of experiments. One restriction is that we can only represent workpieces with right-angled contour elements. For example, a conical contour cannot be represented. Many different cutting and chucking tools are available in real-life process planning. We have restricted ourselves to a single chucking tool and three different cutting tools. The specification of the tools themselves have also been simplified. For example, the rotation speed of workpiece and the feed of the cutting tool are also parameters that can play a role when processing a workpiece. The impact of these parameters has also been neglected. Despite these simplifications the remaining part of this real-world domain is not trivial and represents a substantial subset of the most critical problems in this domain.

---

13. This domain was adapted from the CaPlan-System (Paulokat & Wess, 1994), developed at the University of Kaiserslautern.

14. Note that this figure shows a 2-dimensional drawing of the 3-dimensional workpiece. The measure 1 in. equals 25.4 mm.

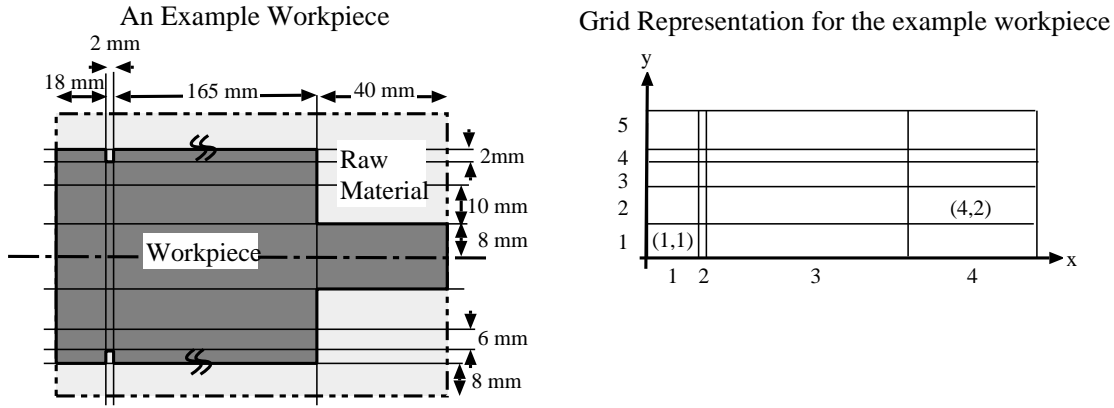

Figure 9: An example workpieces with grid representation

## 8.1 Concrete Domain

We now explain the concrete problem solving domain by giving a detailed description of the states and the operators.

### 8.1.1 STATE DESCRIPTION

For the representation of this domain at the concrete level, the exact geometry of the workpiece must be represented as a state, including the specific measures of each detail of the contour. However, the complete workpiece can always be divided into atomic areas which are always processed as a whole. Therefore the state representation is organized by using a grid which divides the entire workpiece into several disjoint rectangular areas of different sizes (see the right side of Figure 9). Together with a grid coordinate the specific position and size of the corresponding rectangular area are represented. This grid is used as a static part of the state description which does not change during planning. However different problems require different grids. The specific shape of a workpiece during planning is represented by specifying the status for each grid rectangle. In Table 1 the predicates used to represent the workpiece are described in more detail.

Besides the description of the workpiece, the state representation also contains information about how the workpiece is chucked and which kind of cutting tool is currently used. Table 2 describes the predicates which are used for this purpose.

| Predicate | Description |
|---|---|
| *xpos_max* *ypos_max* | The predicates *xpos_max($x_{grid}$)* and *ypos_max($y_{grid}$)* specify the size of the grid in the direction of the x-coordinate and the y-coordinate respectively. A state consists of exactly one instance of each of these predicates, e.g., *xpos_max(4)* and *ypos_max(5)* in the example shown in Figure 9. |
| *grid_xpos* *grid_ypos* | The predicates *grid_xpos($x_{grid}, x_{start}, x_{size}$)* and *grid_ypos($y_{grid}, y_{start}, y_{size}$)* specify the geometrical position and size of grid areas in the direction of the x-coordinate and y-coordinate respectively. The first argument of these predicates specifies the coordinate of the grid areas, the second argument declares the geometrical starting position, and the third argument specifies the size of the grid areas. A state consists of exactly one instance of each of these predicates for each different x-coordinate and y-coordinate. For the example above, *grid_xpos(1,0,18)*, *grid_xpos(2,18,2)*, *grid_xpos(3,20,165)*, *grid_xpos(4,185,40)* specify the grid in x-direction and *grid_ypos(1,0,8)*, ..., *grid_ypos(5,26,8)* specify the grid in y-direction. |
| *mat* | The predicate *mat($x_{grid}, y_{grid}, status$)* describes the status of a particular grid area specified by the coordinates ($x_{grid}, y_{grid}$). The argument *status* can be instantiated with one of the three constants *raw*, *workpiece*, or *none*. The constant *raw* indicates that the specified area still consists of raw material which must be removed by further cutting operators. The constant *workpiece* specifies that the area consists of material that belongs to the goal workpiece. The constant *none* specifies that the area does not contain any material, i.e., there was no material present in the mold or the material has already been removed by previous cutting operations. One instance of a *mat* predicate is required for each grid area to specify its current state. While the previously mentioned predicates does not change during the execution of a plan, the *mat* predicate is changed by each cutting operator. In particular, the initial state and the goal state of a problem differs in the status assigned to those grid areas that must become removed. For example, in the initial state of the example shown above, the sentence *mat(4,2,raw)* will be present while the final state contains the sentence *mat(4,2,none)*. |

Table 1: Essential sentences for the representation of the workpiece

### 8.1.2 Operators

A process plan to manufacture a certain workpiece consists of a sequence of operators. The total order of the operators is not a problem for this domain because the manufacturing steps are also executed sequentially on a lathe.[15] We have chosen four different operators to represent the chucking of a workpiece, the selection of a cutting tool, and the cutting process itself. These operators are described in Table 3.

---

15. However, there are also a few new brands of lathe machine which also allow parallel processing.

| Predicate | Description |
|---|---|
| *chuck_pos* | The predicate *chuck_pos(side)* describes whether the workpiece is currently chucked on either side. The parameter *side* can be instantiated with one of the three constants *none*, *right*, or *left*. The constant *none* specifies that the workpiece is not chucked at all and the constants *right* and *left* specify that the workpiece is chucked at the respective side. Each state contains exactly one instance of this predicate. |
| *covered* | The predicate *covered($x_{min}, x_{max}$)* specifies the areas of the workpiece which are currently covered by the chucking tool. This predicate declares those areas with an x-coordinate lying within the interval $[x_{min}, x_{max}]$ as being covered. Covered areas cannot be processed by a cutting tool. A state consist of exactly one instance of this predicate if the workpiece is chucked. |
| *cut_tool* *cut_direction* | The predicates *cut_tool(id)* and *cut_direction(dir)* specify a unique identification (*id*) of the cutting tool which is currently used when an area is processed and the direction (*dir*) in which the cutting tool moves. The parameter *id* can be any symbol that specifies a legal cutting tool described by predicates included in the static rules $\mathcal{R}_c$ of the concrete domain description. The parameter *dir* can be instantiated by one of the three constants *left*, *right* and *center*. The value *left* specifies that the cutting tool moves from left to right, *right* specifies that the cutting tool moves from right to left, and *center* specifies that the cutting tool move from outside towards the center of the workpiece. |

Table 2: Essential sentences for the representation of the chucking and cutting tools

Manufacturing the workpiece shown in Figure 9 requires a 15-step plan as shown in Figure 10. At first, the workpiece is chucked on the left side. Then a cutting tool is selected which allows cutting from right to left. With this tool the indicated grid areas are removed. Please note that the left side of the workpiece cannot be processed since it is covered by the chucking tool. Then (see the right side of Figure 10), the workpiece is unchucked and chucked on its right side. With a tool that allows processing from left to right, the upper part of the mold is removed. Finally, a specific tool is used to manufacture the small groove.

## 8.2 Abstract Domain

In this example we can see that the small groove can be considered a detail which can be processed after the basic contour of the workpiece has been established. The most important characteristic of this example is that the right part of the workpiece is processed before the left side of the workpiece. This sequence is crucial to the success of the plan. If the groove would have been processed first the workpiece could never be chucked on the left side and the processing of the right side would consequently be impossible. Domain experts told us that this situation is not specific for the example shown. It is of general importance for many cases. This fact allows us to select parts of the problem description and the solution which can be considered as details from which we can abstract. Parts which are "essential" must be maintained in an abstract case. We found out that we can abstract from the detailed shape of the workpiece as long as we distinguish between the processing of the left and right side of the workpiece. Furthermore, it is important to distinguish between the rough contour of the workpiece and the small details such as grooves. We have developed an abstract domain definition containing a new language for describing states and operators based on this abstraction idea.

| Operator | Description |
|----------|-------------|
| *chuck* | The operator *chuck(side)* specifies that the workpiece is chucked at the specified side. The *side* parameter can be instantiated with the constants *left* and *right*. Chucking is only allowed if the workpiece is not chucked already and if the surface used for chucking is plain. As effect of the chucking operation, respective instances of the predicate *chuck_pos* and *covered* are included in the state description. |
| *unchuck* | The operator *unchuck* specifies that the chucking of the workpiece is removed. This operation can only be executed if the workpiece is chucked already. As effect of this operation, the parameter of the predicate *chuck_pos* is changed to *none* and the predicate *covered* is deleted. |
| *use_tool* | The operator *use_tool(dir, id)* specifies which tool is selected for the subsequent cutting operators and in which direction the cutting tool moves. The workpiece must be chucked before a tool can be chosen. The effect of the operator is that respective instantiations of the predicates *cut_tool* and *cut_direction* are added to the state. The parameters of the *use_tool* operator have the same definition as in the respective predicates. |
| *cut* | The operator *cut($x_{grid}, y_{grid}$)* specifies that the raw material in the grid area indicated by the coordinates ($x_{grid}, y_{grid}$) is removed. The effect of this operator is that the predicate *mat* which specifies the status of this particular area is changed from status *raw* to the status *none*. However, to apply this operator several preconditions must be fulfilled. The workpiece must be chucked and the chucking tool must not cover the specified area and the area must be accessible by the cutting tool. Moreover, a cutting tool which allows the processing of the selected area must already have been selected. Each cutting tool imposes certain constraints on the geometrical size of the area that can be processed with it. For details, see the full description of the domain in Online Appendix 1. |

Table 3: Concrete operators

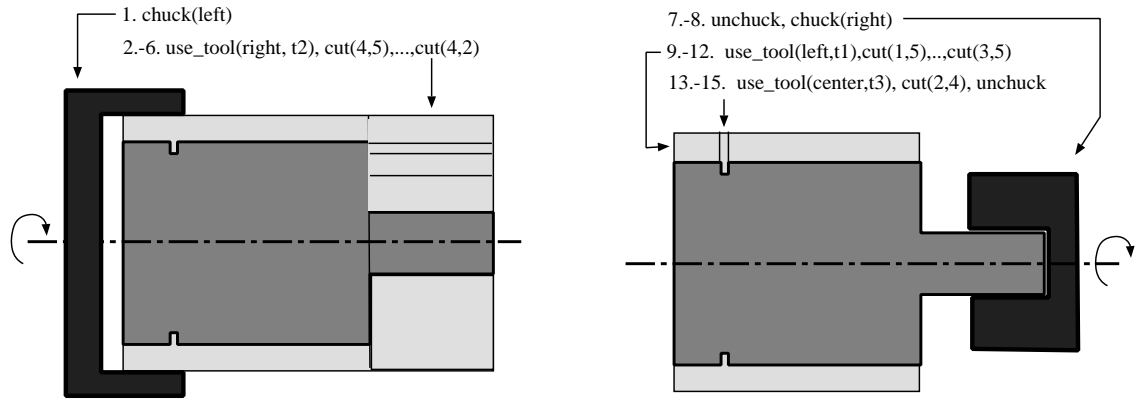

Figure 10: A plan for manufacturing the workpiece

### 8.2.1 STATE DESCRIPTION

We introduce a new abstract grid which divides the workpiece into a left, a middle, and a right area to abstract from the specific location of a concrete grid area. These areas are called complex processing areas. Each area is assigned a particular status. Furthermore, an abstract state contains the information whether a complex processing area contains small contour elements (such as grooves), but not how these grooves exactly look like. To abstract from the very detailed conditions for chucking a workpiece, an abstract state only contains an approximation of these conditions, stating that a workpiece cannot be chucked at a particular side, if this side contains small contour elements that have been already processed. The predicates used to represent an abstract state are described in more detail in Table 4.

### 8.2.2 OPERATORS

We consider an abstract operator which completely processes one complex area of the workpiece, an operator which only processes a complex area roughly, and an operator which processes all the small grooves of a complex area. We also consider an abstract chucking operator because the chucking has a strong impact on the overall plan. Table 5 shows the available abstract operators.

## 8.3 Generic Abstraction Theory

The generic abstraction theory defines the sentences used to describe an abstract state (see Table 4) in terms of the sentences of the concrete state (see Tables 1 and 2) by a set of Horn rules. The definition of abstract sentence is explained in more detail in Table 6.

| Predicate | Description |
|---|---|
| *abs_area_state* | The predicate *abs_area_state(area, status)* describes the status of each of the three complex processing areas. The argument *area* specifies one of the complex processing areas *left*, *middle*, and *right*. The argument *status* describes the status of the respective area. The status can be either *todo*, *rough*, and *ready*. The status *todo* specifies that the area needs some processing of large contour elements, while in a *rough* area only some small contour elements such as grooves need to be processed. The status *ready* specifies that the area is completed. An abstract initial state usually contains one or more complex processing areas of the status *todo*, while in the abstract goal state all complex processing areas have the status *ready*. |
| *abs_small_parts* | The predicate *abs_small_parts(area)* specifies that the complex processing area (*area*) contains small contour elements that need to be manufactured. |
| *abs_chuck_pos* | The predicate *abs_chuck_pos(side)* describes whether the workpiece is currently chucked on either side. The parameter *side* can be instantiated with one of the three constants *none*, *right*, or *left*. This predicate has exactly the same meaning as the *chuck_pos* predicate at the concrete level. This predicate is not abstracted at all but only renamed. |
| *abs_chuckable_wp* | The predicate *abs_chuckable_wp(side)* describes whether the workpiece can be chucked at the *left* or *right* side if this side has been completely processed. |

Table 4: Essential sentences for describing an abstract state

We have strongly considered the factors that influence the quality of a domain (see Section 7.5) during the development of the abstract problem solving domain and the generic abstraction theory. Although none of the defined abstract operators is independently refinable, all of them are *mostly* independently refinable. The preconditions of each abstract operator still contains approximations of the conditions that must be fulfilled in order to assure that a concrete operator sequence exist that refines the abstract operator. For example, the predicate *abs_chuckable_wp(side)* is an approximation of the detailed condition (a plain surface) required for chucking. The goal distance of each operator is quite different and strongly depends on the problem to be solved. While the goal distance of the *set_fixation* operators is no more than two (possibly one *unchuck* operator followed by a *chuck* operator) the goal distances of the other abstract operators are different. For example, the goal distance of the *process_ready* operator depends on the number of concrete grid areas belonging to the respective abstract area and containing material that needs to be removed. The goal distance is the number of these gird areas, say $c$, plus the number of required *use_tool* operations (less than or equal to $c$). Hence, the goal distance is between $c$ and $2c$. Because this goal distance can become very long for the more complex problems, the two operators *process_rough* and *process_fine* are introduced. They only cover the processing of the small and the large grid areas respectively and consequently have a smaller goal distance than the *process_ready* operator. While the goal distance of these two operators is smaller they have a smaller concrete scope of applicability than the *process_ready* operator. For example the *process_ready* operator can be applied in any state in which some arbitrary areas need to be processed, but *process_fine* can only be applied in states in which all large grid areas are already processed.

| Operator | Description |
|----------|-------------|
| *set_fixation* | The operator *set_fixation(side)* specifies that the workpiece is chucked at the specified side. The side parameter can be instantiated with the constants *left*, *right* and *none*. The constant *none* specifies that the chucking is removed. Compared to the concrete operator *chuck* the preconditions for chucking at a side have been simplified. The effect of this operator is that the predicate *abs_chuck_pos* is modified. |
| *process_rough* | The operator *process_rough(area)* specifies that the complex processing area (*area*) is being processed completely up to the small contour elements. The parameter *area* can be either *left*, *middle*, or *right*. The precondition of this operator only requires that the workpiece is chucked at a different side than *area*. The effect of this operator is that the predicate *abs_area_state* is modified. |
| *process_fine* | The operator *process_fine(area)* specifies that all small contour elements of the complex processing area (*area*) are being processed. The parameter *area* can be either *left*, *middle*, or *right*. The precondition of this operator only requires that the large contour elements of this side of the workpiece are already processed and that the workpiece is chucked at a different side. The effect of this operator is that the predicate *abs_area_state* is modified. |
| *process_ready* | The operator *process_ready(area)* specifies that the indicated complex area of the workpiece is being completely processed, including large and small contour elements. The effect of this operator is that the predicate *abs_area_state* is modified. |

Table 5: Abstract operators

Although we have only developed a simplified version of the whole domain of production planning in mechanical engineering for rotary symmetrical workpieces we feel that a domain expert together with a knowledge engineer will be able to define an abstract domain representation and a generic abstraction theory for a complete domain. In particular, model-based interactive knowledge acquisition tools like MIKADO (Schmidt, 1994; Schmidt & Zickwolff, 1992) can make such a complete modeling task much more feasible.

| Abstract Predicate | Description in terms of the predicates of the concrete domain |
|---|---|
| *abs_area_state* | The predicate *abs_area_state(area, status)* describes the status of each of the three complex processing areas. The left processing area consists of the areas of the concrete grid which are covered, if the workpiece is chucked at the left side. Similarly, the right processing area consists of those concrete grid areas which are covered if the workpiece is chucked at the right side. The middle processing area consists of those areas which are never covered by any chucking tool. The status of a complex processing area is *todo*, if there exists a concrete large grid area which belongs to the complex processing area and which needs to be processed. A grid area is considered as large if its size in direction of the x-coordinate is larger than 3 mm. The status of a complex processing area is *rough*, if all large grid areas of the complex processing area are already processed and if there exists a concrete small grid area which belongs to the complex processing area and which needs to be processed. A gird area is considered as small if its size in direction of the x-coordinate is smaller or equal than 3 mm. The status of a complex processing area is *ready* if all concrete grid areas which belong to the complex processing area have been processed. |
| *abs_small_parts* | The sentence *abs_small_parts(area)* holds if there exists a small concrete grid area (size smaller or equal than 3 mm) which belongs to the complex processing area and which needs to be processed. |
| *abs_chuck_pos* | The sentence *abs_chuck_pos(side)* holds if and only if the concrete sentence *chuck_pos(side)* holds. |
| *abs_chuckable_wp* | The predicate *abs_chuckable_wp(side)* describes whether the workpiece can still be chucked at the *left* or *right* side if this side is completely processed. This sentence holds if the part of the desired workpiece which belongs to respective side is completely plain. That is, all concrete grid areas with the status *workpiece* range up to the same y-coordinate. |

Table 6: Generic abstraction theory

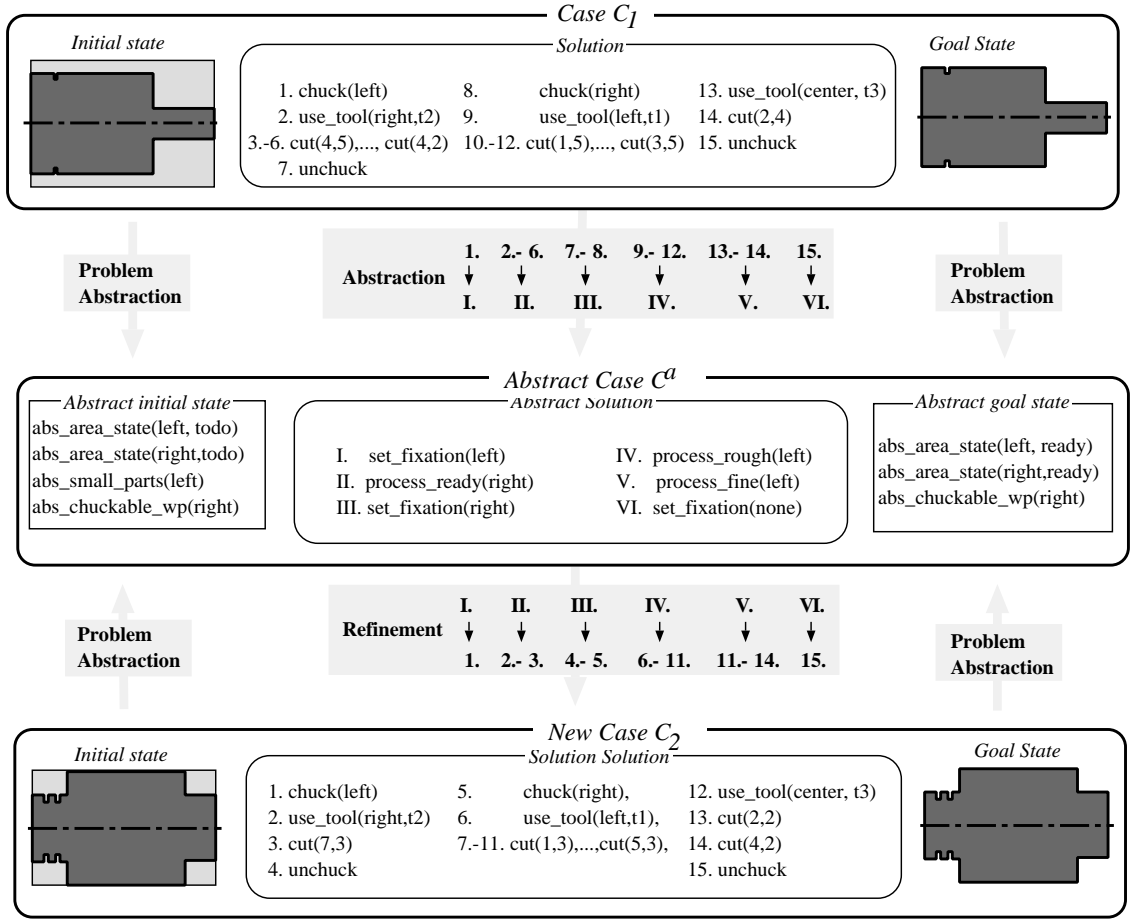

Figure 11: Abstracting and Refining an Example Case

## 8.4 Abstracting and Refining a Process Planning Case

We now explain how the example case shown in Figure 9 can be abstracted and how this abstract case can be reused to solve a different planning problem. This process is demonstrated in Figure 11. The top of this figure shows the concrete planning case $C_1$ already presented in Figure 9. This case is abstracted by the PABS algorithm presented in Section 6. The algorithm returns 6 different abstract cases[16]. One of these abstract cases is shown in the center of the figure. The abstract solution plan consists of a sequence of 6 abstract operators. The sequence of the operators in the plan is indicated by the Roman numerals. The particular abstraction is indicated between the concrete and the abstract case and denotes which sequence of concrete operators is turned into which abstract operator.

---

16. The other 5 abstract cases differ from the shown abstract case in two aspects: In the shown abstract solution the additional abstract step *set_fixation(none)* can be inserted between the steps II and III. The abstract step V can also be replaced by the abstract step *process_ready*, or the abstract steps IV and V together can be replaced by the abstract step *process_ready*.

The learned abstract case can now be used to solve the new problem $C_2$ whose initial and final concrete states are shown in the bottom of the figure. Even if the concrete workpiece looks quite different from the workpiece in case $C_1$ the abstract case can be used to solve the problem. The reason for this is that the new workpiece also requires that the left and right side must be processed. In particular the right side must also be processed before the left side is processed because the left side contains two small grooves which prevent the workpiece from being be chucked at that side after it is processed. However, we can see that most abstract operators (in particular the operators II, VI, and V) are refined to completely different sequences of concrete operators than those from which they were abstracted.

As already mentioned, the abstract operators used are not independently refinable but only *mostly* independently refinable. Consequently, it can happen that an applicable abstract case cannot be refined. Figure 12 shows an example of a concrete planning problem for which the abstract case shown in Figure 11 is applicable but not refinable. The reason for this is the location of the small abstract part at the left side of the workpiece. This small part consists of the concrete grid area (1,3) in which raw material must be removed. However, in this specific situation, this small part must be removed *before* the large parts, the left side of the workpiece contains (the grid areas (2,3), (3,3), and (2,2)), can be removed. The reason for this is that without removing this small part, the larger parts located right of the small part cannot be accessed by any cutting tool that is able to cut the areas (2,3) and (3,3). Consequently this problem can only be solved with the plan shown on the right side of Figure 12. Unfortunately, this plan is not a refinement of the abstract plan shown in Figure 11, because this abstract plans requires that the large parts must be removed before the small parts are removed. Hence, the refinement of the operator *process_rough(left)* fails. In this situation the problem solver must select a different abstract plan.

## 9. Empirical Evaluation and Results

This section presents the results of an empirical study of Paris in the mechanical engineering domain already introduced. This evaluation was performed with the fully implemented Paris system using only the abstraction abilities of the system. The generalization component was switched-off for this purpose. We have designed experiments which allow us to judge the performance improvements caused by various abstract cases derived by Pabs. Furthermore, we have analyzed the average speed-up behavior of the system with respect to a large set of randomly selected training and test cases.

### 9.1 Planning Cases

For this empirical evaluation 100 concrete cases have been randomly generated. Each case requires about 100-300 sentences to describe the initial or final state, most of which are instances of the *mat* predicate. The length of the solution plans ranges from 6 to 18 operators. Even if the generated cases only represent simple problems compared to the problems a real domain expert needs to solve, the search space required to solve our sample problems is already quite large. This is due to the fact that the branching factor $b$ is between 1.7 and 6.6, depending on the complexity of the problem. Hence, for a 18-step solution the complete search space consists of $3.7 \cdot 10^{15}$ states.

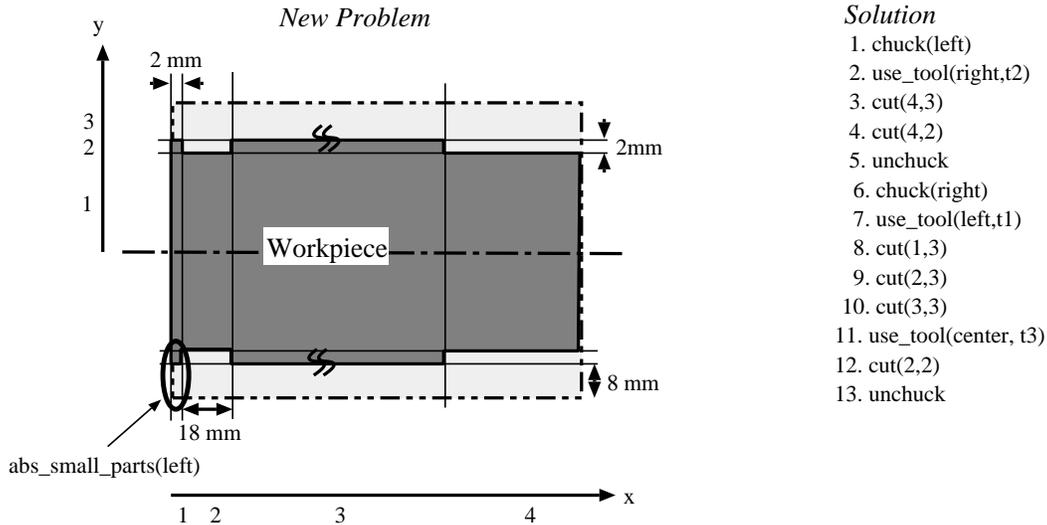

Figure 12: An Example Case in which the refinement of the abstract plan shown in Figure 11 fails.

The case generation procedure leads to solutions which are optimal or nearly optimal. All solutions which require less than 10 steps are optimal solutions in the sense that they are known to be the shortest solution to the problem they solve. All solutions which are longer than 10 steps have been manually checked to see whether they contain steps which are obviously redundant. Such redundant steps have been removed. Although these solutions are not necessarily shortest solutions, they are nevertheless acceptably short.

## 9.2 Evaluating Abstraction by Dropping Sentences

At first we used the recent version of ALPINE (Knoblock, 1993) together with PRODIGY-4 (Blythe et al., 1992) to check whether abstraction by dropping sentences can improve problem solving in our domain represented as described in Section 8. Therefore, we used only the concrete problem solving domain as domain theory for PRODIGY. Unfortunately, for this representation, ALPINE was not able to generate an ordered monotonic abstraction hierarchy. The reason for this is that ALPINE can only distinguish a few different groups of literals because only a few different literal names (and argument types) can be used in the problem space. For example, ALPINE cannot distinguish between the different sentences which are described by the *mat* or the *grid_xpos* predicate. But this is very important for abstraction. We would like to drop those parts of the grid which represent small rectangles such as grooves. However, this would require the examination of the measures associated with a grid area (as argument) and also the relation to other surrounding grid areas. Therefore, which sentence to drop (or which criticalities to assign) cannot be decided statically by the name of the predicate or the type of the arguments. All hierarchical planners including

Prodigy and Alpine are highly dependent on the representation used, in particular if their strategy is restricted to dropping sentences (Holte et al., 1994, 1995). However, there might be another representation of our domain for which those hierarchical planners can improve performance but we think that our representation is quite "natural" for our domain.

From this first trial we can conclude that the application domain and representation we have chosen for the following experiments with Paris really require more than dropping sentences to achieve an improvement by abstraction.

## 9.3 Evaluating the PARIS Approach

The first experiment with Paris was designed to evaluate the hypotheses that in our domain there is a need (I) for changing the representation language during abstraction, and (II) for reusing abstract cases instead of generating abstract solutions from scratch. To test these hypotheses we rely on the time for solving the randomly generated problems using different modes of the Paris system.

### 9.3.1 Experimental Setting

In this experiment we used the Paris system to solve the 100 problems from the randomly generated cases. Thereby the goal of abstraction is to improve the concrete-level problem solver, which performs a brute-force search with a depth-first iterative-deepening search strategy (Korf, 1985a) as introduced in Section 7.3. The improvement is determined in terms of problem solving time required to solve a single problem. Paris is used to solve the 100 problems in three different modes:

- Pure search: The problem solver is used to solve each problem by pure search without use of any abstraction.

- Hierarchical planning: In this mode Paris uses the introduced abstract domain. However, abstract cases are not recalled from a case library but they are computed automatically by search as in standard hierarchical planning, but using the new abstraction language. So, the problem solver first tries to search for a solution to the original problem at the abstract domain and then tries to refine this solution. During this hierarchical problem solving, backtracking between the two levels of abstraction and between each subproblem can occur. Thereby, we used hierarchical planning with the new abstraction methodology instead of dropping sentences.

- Reasoning from abstract cases: In this mode we first used Paris to learn all abstract cases which come out of the 100 concrete cases. For each problem, all abstract cases that exists according to our abstraction methodology are available when one of the problems is to be solved. During problem solving we measured the time required for solving *each* problem using *every* applicable abstract cases. Then, for each problem, three abstract cases are determined: a) the best abstract case, i.e., the case which leads to the shortest solution time, b) the worst abstract case (longest solution time) which is an abstraction of the aspired solution case, and c) the worst *applicable* abstract case is determined. The difference between b) and c) relates to the difference between applicable and refinable abstract cases introduced in Section 7.1. An abstract case selected in c) is applicable to the current problem, but might not be an abstraction of the case from which the problem is taken. In b) only abstract cases are selected which are indeed abstractions of the current problem, i.e., abstract cases which have been previously learned from the case from which the problem is taken. These three different cases are selected to figure out the impact of case selection (which is not addressed in this paper) on the proposed method.

Although every problem can theoretically be solved by our brute-force search procedure, the exponential nature of the search space avoids the solution of complex problems within reasonable time. Therefore, a time-bound of 200 CPU seconds on a SUN SPARC-ELC computer was introduced in each of the three modes described above. If this limit-bound is exceeded the problem remains unsolved. Increasing this time-bound would increase the number of solvable problems in each of the three modes.

### 9.3.2 RESULTS

We have determined the solution time for each of the 100 problems in each of the described modes. The average solution time as well as the number of problems that could be solved within the time limit is shown in Table 7. We have determined these values for reasoning from abstract cases separately for each of the three types of abstract cases. The significance of the speedup results has be investigated by using a maximally conservative sign test (Etzioni & Etzioni, 1994). Unfortunately it turned out that the speedup of hierarchical planning over pure search was not significant. We also couldn't find a significant speedup of reasoning from abstract cases when using always the worst applicable abstract case (c) over pure search. This was due to the large number of doubly censored data (both problem solvers cannot solve the problem within the time limit), which were counted against the speedup hypothesis. However, the improvements of pure search by reasoning from refinable abstract cases were significant ($p < 0.000001$) when using the best refinable case (a) and when using the worst refinable case (b). Furthermore, it turned out that the speedup of reasoning from refinable cases over hierarchical planning was also significant for an upper bound of the p-value of 0.001. The mentioned *p-value* is a standard value used in statistical hypothesis tests. It is the probability, assuming that the hypothesis does not hold, of encountering data that favors the hypothesis as much or more than the observed data in the experiment (Etzioni & Etzioni, 1994). Therefore a result is more significant if the p-value is smaller. From this analysis, we can clearly see, that our two basic hypotheses are supported by our experimental data. Even if not significant we can see a moderate improvement in the problem solving time and in the number of solved problems when using hierarchical planning with changing the representation language. Please remember that hierarchical planning by dropping conditions did not lead to any improvement at all (see Section 9.2). Obviously, changing the representation language during abstraction is required to improve problem solving in our domain as stated in the first hypothesis (I).

Very strong support for the second hypothesis (II) can also be found in the presented data. We can see significant speedups by reasoning from abstract cases over pure search and even over hierarchical planning. Only if the worst abstract case is used for each problem to be solved, the speedup is not significant and the problem solving behavior is slightly worse than in hierarchical planning. Please note that this situations is extremely unlikely to happen at all. With a sophisticated indexing and retrieval of abstract cases this situation can be avoided for the most part.

| Problem solving mode | Average solution time (sec.) | Solved problems |
|---|---|---|
| Pure search | 156 | 29 |
| Hierarchical planning | 107 | 50 |
| Reasoning from abstract cases | | |
| (a) Best refinable case | 35 | 94 |
| (b) Worst refinable case | 63 | 79 |
| (c) Worst applicable case | 117 | 45 |

Table 7: Comparison of the average solution time per problem and the number of solved problems within a time-bound of 200 seconds. The table compares pure search (depth-first iterative deepening), hierarchical planning using the abstract problem solving domain, and reasoning from abstract cases with differently selected abstract cases.

## 9.4 Evaluating the Impact of Different Training Sets

In one respect the previous experiment is based on a very optimistic assumption. We always assume that all abstract cases required for solving a problem have been learned in advance. This situation is not a realistic scenario for an application. Usually, one set of cases is available for training the system while a different set of problems needs to be solved. So we cannot assume that good applicable abstract cases are always available to solve a new problem. Furthermore, the presented example also shows that the problem solving time can vary a lot if different abstract cases are selected during problem solving. Therefore, we have designed a new experiment to evaluate the improvements caused by the PARIS approach in a more realistic scenario.

### 9.4.1 Experimental Setting

We have randomly chosen 10 training sets of 5 cases and 10 training sets of 10 cases from the 100 available cases. These training sets are selected independently from each other. Then, each of the 20 training sets is used for a separate experiment. In each of the 20 experiments, those of the 100 cases which are not used in the particular training set are used to evaluate the performance of the resulting system. Training set and test set are completely independent by this procedure. During this problem solving task, we did not determine the problem solving behavior for *all* applicable abstract cases, but we used a simple automatic mechanism to retrieve *one* (hopefully a good) applicable abstract case for a problem. Therefore, the cases are organized linearly in the cases base, sorted by the length of the abstract plan contained in the case. The case base is sequentially searched from longer to shorter plans until an applicable case is found. This heuristic is based on the assumption that a longer abstract plan is more specific than a shorter abstract plan and divides the actual problem into more, but smaller subproblems. Consequently the longest applicable plan should lead to the best improvement.

| Size of training sets | Number of abstract cases | | |
|---|---|---|---|
| (cases) | minimum | maximum | average |
| 5 | 7 | 15 | 9.1 |
| 10 | 8 | 25 | 14.2 |

Table 8: Comparison of the number of learned abstract cases for a) the 10 training sets each of which consists of 5 concrete cases and b) the 10 training sets each of which consists of 10 concrete cases. The table shows the minimum, the maximum, and the average number of abstract cases learned from the 10 training sets of the respective size.

| Size of training sets | Average problem solving time (sec.) | | |
|---|---|---|---|
| (cases) | best set | worst set | average |
| 5 | 43 | 89 | 59 |
| 10 | 35 | 76 | 56 |

Table 9: Comparison of the problem solving time required for reasoning from abstract cases after separate training with a) the 10 training sets each of which consists of 5 concrete cases and b) the 10 training sets each of which consists of 10 concrete cases. The table shows the average problem solving time per problem for the best, the worst and the average training set out of the 10 training sets of each size.

### 9.4.2 RESULTS

We have statistically evaluated the second experiment. Table 8 shows the number of abstract cases which could be learned from the different training sets. The minimum, the maximum and the average number of abstract cases that could be learned from the 10 training sets of the same size is indicated. Note that altogether 42 abstract cases can be learned if all 100 cases would have been used for training as in the previous experiment. From the 10 training sets which contained 5 cases each, between 7 and 15 abstract cases could be learned. As expected, if the size of the training set is increased more abstract cases can be learned. Table 9 shows the average problem solving time after learning from the different sets. This table also shows the minimum, the maximum and the average problem solving time for the 10 different training sets of the two sizes. We can see that the best training sets leads to a problem solving time which is similar or only slightly worse than the optimum shown in Table 7. Even in the average case, considerable improvements over the pure search and hierarchical problem solving (compare Table 7 and Table 9) can be discovered. The same positive results can also be identified when looking at the percentage of solved problems, shown in Table 10. Here we can also see that for the best training sets the number of solved problems is close to the maximum that can be achieved by this approach. Even in the worst training set considerably more problems could be solved than by pure search or hierarchical planning.

| Size of training sets | Percentage of Solved Problems | | |
|---|---|---|---|
| (cases) | best set | worst set | average |
| 5 | 91 | 68 | 83 |
| 10 | 94 | 74 | 86 |

Table 10: Comparison of the percentage of solved problems after separate training with a) the 10 training sets each of which consists of 5 concrete cases and b) the 10 training sets each of which consists of 10 concrete cases. The table shows the percentage of solved problems for the best, the worst and the average training set out of the 10 training sets of each size.

| Size of training sets | Number of training sets with significant speedups over | | |
|---|---|---|---|
| (cases) | pure search | hierarchical planning | |
| | $p < 0.0005$ | $p < 0.0005$ | $p < 0.05$ |
| 5 | 9 | 4 | 8 |
| 10 | 10 | 5 | 7 |

Table 11: Comparison of the significance (p-value) of the speedup results over pure search and hierarchical planning after separate training with a) the 10 training sets each of which consists of 5 concrete cases and b) the 10 training sets each of which consists of 10 concrete cases. The table shows the number of training sets which cause significant speedups for different p-values.

Additionally all of the above mentioned speedup results were analyzed with the maximally conservative sign test as described in (Etzioni & Etzioni, 1994). Table 11 summarizes the significance results for speeding up pure search and a hierarchical problem solver. It turned out that 19 of the 20 training sets lead to highly significant speedups ($p < 0.0005$) over pure search. For this hard upper bound on p-values only about half of the training sets lead to significant differences between reasoning from abstract cases and hierarchical planning. At a slightly higher upper bound of $p < 0.05$, about 3/4 of the training sets caused a significantly better performance than hierarchical planning.

Altogether, the reported experiment showed that even a small number of training cases (i.e., 5% and 10%) can already lead to strong improvements on problem solving. We can see that not all abstract cases must be present, as in the first experiment, to be successful. Furthermore, this experiment has shown that even a simple retrieval mechanism (sequential search) can select beneficial abstract cases from the library. Neither of the training situations in the second experiment lead to results which are as worse as the worst case shown in Table 7.

| Size of training sets | Average percentage of solutions with shorter/equal/longer solution length | | |
|:---:|:---:|:---:|:---:|
| (cases) | shorter | equal | longer |
| 5 | 20 | 54 | 26 |
| 10 | 22 | 50 | 28 |

Table 12: Comparison of the length of the solutions created through reasoning from learned abstract cases and the solutions available in the concrete cases. The table shows the average percentage of solutions with shorter/equal/longer solution length after separate training with a) the 10 training sets each of which consists of 5 concrete cases and b) the 10 training sets each of which consists of 10 concrete cases.

## 9.5 Quality of the Produced Solutions

Although the main purpose of this approach is to improve the performance of a problem solver, the quality of the produced solutions is also very important for a practical system. The solution length can be used as a very simple criterion to determine the quality of a solution. However, in general the quality of a solution should reflect the execution costs of a plan, the plans robustness, or certain user preferences (Perez & Carbonell, 1993). Because such quality measures are very difficult to assess, in particular in our manufacturing domain, we rely on this simple criterion also used for evaluating the quality of solutions in PRODIGY/ANALOGY (Veloso, 1992).

### 9.5.1 EXPERIMENTAL SETTING

We have analyzed the solutions computed in the previous set of experiments to assess the quality of the solutions produced by PARIS. Therefore, the length of solutions derived during problem solving, after learning from each of the 20 training sets, are compared to the length of the nearly optimal solutions contained in the concrete cases.

### 9.5.2 RESULTS

For each training set the length of each solution derived in the corresponding testing phase is compared to the length of the solution noted in the concrete case. The percentage of solutions with shorter, equal, or longer solution length is determined for each training set separately, and the average over the 10 training sets with equal size is determined. Table 12 shows the result of this evaluation.

It turned out that there was no big difference in the quality results between the 20 training sets. In particular, the size of the training sets did not have a strong influence on the results. In Table 12 we can see that between 72% (22% + 50%) and 74% (20% + 54%) of the solutions produced are of equal or better quality than the solutions contained in the concrete cases. Please note that the concrete cases used for testing are always different from the cases used for training. Additionally, the solutions to which we compare the results produced by PARIS are already nearly optimal solutions due to the case generation procedure.[17] Taking this into account, these results are already fairly good.

## 9.6 Impact of the Abstract Problem Solving Domain

The experiments reported before were conducted with the concrete and abstract domain representation presented in Section 8 and in Online Appendix 1. In this final experiment the impact of the specific choice of an abstract problem solving domain is investigated.

### 9.6.1 EXPERIMENTAL SETTING

We created a new abstract problem solving domain which is less constrained than the one used before. For this purpose one operator was completely removed and certain conditions of the remaining operators were removed also. In particular, the *set_fixation* operator was removed and the conditions *abs_chuck_pos*, *abs_chuckable_wp*, and *chuck_comp* were removed from the preconditions of the three remaining operators. Hence, the fact that the chucking of a workpiece has an impact on the production plan is now neglected at the abstract level. However, the concrete problem solving domain and the generic abstraction theory was not modified at all. Consequently, chucking still plays an important role at the concrete level. The set of experiments described in Section 9.4 was repeated with the less constrained abstract problem solving domain but using the same training and testing sets as before.

### 9.6.2 RESULTS

Table 13 and 14 summarize the results of these experiments. Table 13 shows the average problem solving time which occurs after learning from the different training sets. It turns out that for all training sets, learning improves the concrete level problem solver, but that the speedup is much smaller than when using the original abstract problem solving domain (cf. Table 7 and 9). In particular, none of the resulting speedups over concrete level problem solving were significant. A similar result can be observed when comparing the percentage of solved problems (see Figure 14). There is still a slight improvement in the number of problems that could be solved after learning but the improvement is much smaller than when using the original abstract problem solving domain (cf. Table 7 and 10).

---

17. In all cases up to one, the shorter solutions produced by PARIS are only one step shorter than the solution contained in the concrete case.

| Size of training sets | Average problem solving time (sec.) | | |
|:---:|:---:|:---:|:---:|
| (cases) | best set | worst set | average |
| 5 | 114 | 118 | 117 |
| 10 | 107 | 112 | 110 |

Table 13: Using a less constrained abstract problem solving domain: Comparison of the problem solving time required for reasoning from abstract cases after separate training with a) the 10 training sets each of which consists of 5 concrete cases and b) the 10 training sets each of which consists of 10 concrete cases. The table shows the average problem solving time per problem for the best, the worst and the average training set out of the 10 training sets of each size.

| Size of training sets | Percentage of Solved Problems | | |
|:---:|:---:|:---:|:---:|
| (cases) | best set | worst set | average |
| 5 | 55 | 52 | 53 |
| 10 | 58 | 54 | 56 |

Table 14: Using a less constrained abstract problem solving domain: Comparison of the percentage of solved problems after separate training with a) the 10 training sets each of which consists of 5 concrete cases and b) the 10 training sets each of which consists of 10 concrete cases. The table shows the percentage of solved problems for the best, the worst and the average training set out of the 10 training sets of each size.

This experiment supported the general intuition that the abstract problem solving domain has a significant impact on the improvement in problem solving that can be achieved through reasoning from abstract cases. The reason why the less constrained domain leads to worse results than the original abstract domain can be explained with respect to the criteria explained in Section 7.5. Since important preconditions of the abstract operators were removed there are many situations in which the new operators cannot be refined. This holds particularly for those situations in which a workpiece cannot be chucked to perform the required cutting operations. The new abstract operators are not mostly independently refinable. Moreover, since the abstract operator *set_fixation* is removed the concrete *chuck* and *unchuck* operator must be introduced during the refinement of the remaining abstract operators. Consequently, the goal distance of these abstract operators is increased. These two factors are the reason for worse results when using the less constrained abstract domain theory.

## 10. Discussion

In this paper we have shown in detail that in hierarchical problem solving (Sacerdoti, 1974; Tenenberg, 1988; Unruh & Rosenbloom, 1989; Yang & Tenenberg, 1990; Knoblock, 1990) the limited view of abstraction by dropping sentences as well as the strategy by which abstract solutions are computed lead to poor behavior in various relevant situations. This observation is supported by comprehensive artificial examples (see Section 2.1 and 2.2) and a real-world example from the domain of mechanical engineering (see Section 8), further supported by an experiment (see Section 9.2). The recent results reported in (Holte et al., 1995) support these observations very well.

In general, abstraction is the task of transforming a problem or a solution from a concrete representation into a different abstract representation, while reducing the level of detail (Michalski & Kodratoff, 1990; Giunchiglia & Walsh, 1992; Michalski, 1994). However, in most hierarchical problem solvers, the much more limited view of abstraction by dropping sentences is shown to be the reason why efficient ways of abstracting a problem and a solution are impossible (e.g., see Section 2.1 and Figure 4). The second weakness of most hierarchical problem solvers is that they usually compute arbitrary abstract solutions and not solutions which have a high chance of being refinable at the next concrete level. Although the upward solution property (Tenenberg, 1988) guarantees that a refinable abstract solution exists, it is not guaranteed that the problem solver finds this abstract solution (e.g., see Section 2.2). Problem solvers are not even heuristically guided towards refinable abstract solutions.

With the PARIS approach we present a new formal abstraction methodology for problem solving (see Section 5) which allows abstraction by changing the whole representation language from concrete to abstract. Together with this formal model, a correct and complete learning algorithm for abstracting concrete problem solving cases (see Section 6) is given. The abstract solutions determined by this procedure are useful for solving new concrete problems, because they have a high chance of being refinable.

The detailed experimental evaluation with the fully implemented PARIS system in the domain of mechanical engineering strongly demonstrates that PARIS can significantly improve problem solving in situations in which a hierarchical problem solver using dropping sentences fails to show an advantage (see Table 7 to 11).

### 10.1 Related Work

We now discuss the PARIS approach in relation to other relevant work in the field.

#### 10.1.1 Theory of Abstraction

Within Giunchiglia and Walsh's (1992) theory of abstraction, the PARIS approach can be classified as follows: The formal system of the ground space $\Sigma_1$ is given by the concrete problem solving domain $\mathcal{D}_c$ using the situation calculus (Green, 1969) for representation. The language of the abstract formal system $\Sigma_2$ is given by the language of the abstract problem solving domain $\mathcal{D}_a$. However, the operators of $\mathcal{D}_a$ are not turned into axioms of $\Sigma_2$. Instead, the abstract cases build the axioms of $\Sigma_2$. Moreover, the generic abstraction theory $\mathcal{A}$ defines the abstraction mapping $f : \Sigma_1 \Rightarrow \Sigma_2$. Within this framework, we can view

PARIS as a system which learns useful axioms of the abstract system, by composing several smaller elementary axioms (the operators). However, to prove a formula (the existence of a solution) in the abstract system, exactly one axiom (case) is selected. So the deductive machinery of the abstract system is restricted with respect to the ground space. Depending on the learned abstract cases the abstractions of PARIS are either theory decreasing (TD) or theory increasing (TI). If the case-base of abstract cases is completely empty then no domain axiom is available and the resulting abstractions are consequently TD. If the case-base contains the maximally abstract case $\langle\langle true, true\rangle(nop)\rangle$[18] (and the generic abstraction theory contains the clause $\rightarrow true$), then this case can be applied to every concrete problem and the resulting abstraction is consequently TI. Even if this maximally abstract case does not improve the ground level problem solving, it should be always included into the case-base to ensure the TI property, that is not loosing completeness. The case retrieval mechanism must however guarantee, that this maximally abstract case is only chosen for refinement if no other applicable case is available. Note, that this is fulfilled for the retrieval mechanism (sequential search from longer to shorter plans) we used in our experiments.

### 10.1.2 SKELETAL PLANS

As already mentioned in Section 3.4 the PARIS approach is inspired by the idea of skeletal plans (Friedland & Iwasaki, 1985). A abstract cases can be seen as a skeletal plan, and our learning algorithm is a means to learn skeletal plans automatically out of concrete plans. Even if the idea of skeletal plans is intuitively very appealing, to our knowledge, this paper contains the first comprehensive experimental support of usefulness of planning with skeletal plans. Since we have shown that skeletal plans can be acquired automatically, this planning method can be applied more easily.

For the same purpose, Anderson and Farley (1988) and Kramer and Unger (1992) proposed approaches for plan abstraction which go in the same direction as the PARIS algorithm. However, this approach automatically forms abstract operators by generalization, mostly based on dropping sentences. Moreover, in the abstracted plan, *every* concrete operator is abstracted, so that the number of operators is not reduced during abstraction. Thereby this abstraction approach is less powerful than PARIS style abstractions.

### 10.1.3 ALPINE'S ORDERED MONOTONIC ABSTRACTION HIERARCHIES

ALPINE (Knoblock, 1989, 1990, 1993, 1994) automatically learns hierarchies of abstraction spaces from a given domain description or from a domain description together with a planning problem. As mentioned several times before, ALPINE relies on abstraction by dropping sentences. However, this enables ALPINE to generate abstraction hierarchies automatically. For a stronger abstraction framework such as the one we follow in PARIS, the automatic generation of abstraction hierarchies (or abstract domain descriptions) does not seem to be realistic due to the large (infinite) space of possible abstract spaces. To use our powerful abstraction methodology, we feel that we have to pay the price of losing the ability to automatically construct an abstraction hierarchy.

Another point is that the specific property of ordered monotonic abstraction hierarchies generated by ALPINE, allows an efficient plan refinement. During this refinement, an ab-stract plan can be expanded at successively lower levels by *inserting* operators. Furthermore, already established conditions of the plan are guaranteed not to be violated anymore during refinement. Unfortunately, this kind of refinement cannot be performed for PARIS-style abstractions. Especially, there is no direct correspondence between the abstract operators and concrete operators. Consequently, an abstract plan cannot be *extended* to become a concrete plan. However, the main function of the abstract plan is maintained, namely that the original problem is decomposed into several smaller subproblems which causes the main reduction in search.

---

18. *nop* is the 'no operation' operator which is always applicable and does not change the abstract state.

### 10.1.4 EXPLANATION-BASED LEARNING, CASE-BASED REASONING AND ANALOGY

The presented PARIS approach uses experience to improve problem solving, similar to several approaches from machine learning, mostly from explanation-based learning (Mitchell et al., 1986; DeJong & Mooney, 1986), case-based reasoning (Kolodner, 1980; Schank, 1982; Althoff & Wess, 1992; Kolodner, 1993) or analogical problem solving (Carbonell, 1986; Veloso & Carbonell, 1988). The basic ideas behind explanation-based learning and case-based or analogical reasoning are very much related. The common goal of these approaches is to avoid problem solving from scratch in situations which have already occurred in the past. Explanations (i.e., proofs or justifications) are constructed for successful solutions already known by the system. In explanation-based approaches, these explanations mostly cover the whole problem solving process (Fikes, Hart, & Nilsson, 1972; Mooney, 1988; Kambhampati & Kedar, 1994), but can also relate to to problem solving *chunks* (Rosenbloom & Laird, 1986; Laird, Rosenbloom, & Newell, 1986) of some smaller size or even to single decisions within the problem solving process (Minton, 1988; Minton et al., 1989). Explanation-based approaches *generalize* the constructed explanations during learning by extensive use of the available domain knowledge and store the result in a control rule (Minton, 1988) or schema (Mooney & DeJong, 1985). In case-based reasoning systems like PRIAR (Kambhampati & Hendler, 1992) or PRODIGY/ANALOGY (Veloso & Carbonell, 1993; Veloso, 1994) cases are usually not explicitly generalized in advance. They are kept fully instantiated in a case library, annotated with the created explanations. Unlike cases in PARIS which are problem-solution-pairs, such cases are complete problem solving episodes containing detailed information of each decision that was taken during problem solving. During problem solving, those cases are retrieved which contain explanations applicable to the current problem (Kambhampati & Hendler, 1992; Veloso & Carbonell, 1993; Veloso, 1994). The detailed decisions recorded in these cases are then replayed or modified to become a solution to the current problem. All these approaches use some kind of generalization of experience, but none of these approaches use the idea of abstraction to speedup problem solving based on experience. As already noted in (Michalski & Kodratoff, 1990; Michalski, 1994), abstraction and generalization must not be confused. While generalization transforms a description along a set-superset dimension, abstraction transforms a description along a level-of-detail dimension.

The only exception is given in (Knoblock, Minton, & Etzioni, 1991a) where ALPINE's abstractions are combined with EBL component of PRODIGY. Thereby, control rules are learned which do not refer to the ground space of problem solving but also to the abstract spaces. These control rules speedup problem solving at the abstract level. However, the control rules guide the problem solver at the abstract level so that it finds solutions faster and not in a manner that it finds refinable abstract solutions. Although we did not have any experience with this kind of integration of abstraction and explanation-based learning, we assume that the control rules generated by the EBL component will also guide the problem solver towards short abstract solutions which do not cause much reduction in search in several circumstances.

## 10.2 Requirements and Limitations of PARIS

In the following, we will summarize again the requirements and limitations of the PARIS approach. The main requirements are the availability of a good abstract domain description and in the availability of concrete cases.

### 10.2.1 ABSTRACT DOMAIN

The most important prerequisite of this method is the availability of the required background knowledge, namely the concrete world description, the abstract world description, and the generic abstraction theory. For the construction of a planning system, the concrete world descriptions must be acquired anyway, since they specify the "language" of the problem description (essential sentences) and the problem solution (operators). The abstract world and the generic abstraction theory must also be acquired. We feel that this is indeed the price we have to pay to make planning more tractable in certain practical situations.

Nevertheless, the formulation of an adequate abstract domain theory is crucial to the success of the approach. If those abstract operators are missing which are required to express a useful abstract plan, no speedup can be achieved. What we need are mostly independently refinable abstract operators. If such operators exist, they can be simply represented in the abstract domain using the whole representational power. For hierarchical planning with dropping conditions, such an abstract domain must also be *implicitly* contained in a concrete domain in a way that the abstract domain remains, if certain literals of the concrete domain are removed (see Section 2.1). We feel that this kind of modeling is very much harder to achieve than modeling the abstract view of a domain explicitly in a distinct planning space as in PARIS. Additionally, the requirement that the abstract domain is given by the user has also the advantage that the learned abstract cases are expressed in terms the user is familiar with. Thereby, the user can understand an abstract case very easily. This can open up the additional opportunity to involve the user in the planning process, for example in the selection of an abstract cases she/he favors.

Research on knowledge acquisition has shown that human experts employ a lot of abstract knowledge to cope with the complexity of real-world planning problems. Specific knowledge acquisition tools have been developed to comfortably acquire such abstract knowledge from different sources. Especially, the acquisition of planning operators is addressed in much detail in (Schmidt & Zickwolff, 1992; Schmidt, 1994).

### 10.2.2 AVAILABILITY OF CASES

As a second prerequisite, the PARIS approach needs concrete planning cases (problem-solution pairs). In a real-world scenario such cases are usually available in a company's filing cabinet or database. According to this requirement we share the general view from machine learning that the use of this kind of experience is the most promising way to cope with highly intractable problems. For the PARIS approach the available cases must be somehow representative for future problem solving tasks. The known cases must be similar enough to the new problems that abstract cases can really be reused. Our experiments give strong indications that even a small set of concrete cases for training leads to high improvements in problem solving (see Table 9 to 11).

## 10.3 Generality of the Achieved Results

The reported experiments were performed with a specific base-level problem solver which performs a depth-first iterative-deepening search strategy (Korf, 1985a). However, we strongly believe that the PARIS abstractions are also beneficial for other problem solvers using backward-chaining, means-end analysis or nonlinear partial-order planning. As shown in (Veloso & Blythe, 1994), there is not one optimal planning strategy. Different planning strategies usually rely on different commitments during search. Each strategy can be useful in one domain but may be worse in others. However, for most search strategies, the length of the shortest possible solution usually determines the amount of search which is required. In PARIS, the whole search problem is decomposed into several subproblems which allow short solutions. Consequently, this kind of problem decomposition should be of use for most search strategies.

Moreover, we think that the idea of reasoning from abstract cases, formulated in a completely new terminology than the ground space will also be useful for other kinds of problem solving such as design or model-based diagnosis. For model-based diagnosis, we have developed an approach (Pews & Wess, 1993; Bergmann, Pews, & Wilke, 1994) similar to PARIS. The domain descriptions consist of a model of a technical system for which a diagnosis has to be found. It describes the behavior of each elementary and composed component of the system at different levels of abstraction. During model-based diagnosis, the behavior of the technical system is simulated and a possible faulty component is searched which can cause the observed symptoms. Using abstract cases, this search can be reduced and focused onto components which have been already defective (in other similar machines) and which are consequently more likely to be defective in new situations.

## 10.4 Future Work

Future research will investigate goal-directed procedures for refinement such as backward-directed search or non-linear partial order planners (see Section 7.4). Additionally, more experience must be gained with additional domains and different representations of them. Furthermore, we will address the development of highly efficient retrieval algorithms for abstract cases. As Table 7 shows, the retrieval mechanism has a strong influence on the achieved speedup. Even if the linear retrieval we have presented turned out to be pretty good, we expect a utility problem (Minton, 1990) to occur when the size of the case-base grows. Furthermore, a good selection procedure for abstract cases should also use some feedback from the problem solver to evaluate the learned abstract cases based on the speedup they cause. This would eliminate unbeneficial cases or abstract operators from the case-base or the abstract problem solving domain. Experiments with different indexing and retrieval mechanisms have recently indicated that this is possible.

Furthermore, the speedup caused by a combination of different approaches such as abstraction and explanation-based learning should be addressed. Within the Paris system an explanation-based component for case generalization is still present (see Figure 3), but was not used for the experiments because the plain abstraction itself had to be evaluated. In further experiments, abstraction, explanation-based learning and the integration of both has to be addressed comprehensively. This will hopefully lead to a better understanding of the different strengths these methods have.

As a more long-term research goal, Paris-like approaches should be developed and evaluated for other kinds of problem solving tasks such as configuration and design or, as already started, for model-based diagnosis.

## Appendix A. Proofs

This section contains the proofs of the various lemma and theorems.

**Lemma 6** (Joining different abstractions) *If a concrete domain $\mathcal{D}_c$ and two disjoint abstract domains $\mathcal{D}_{a1}$ and $\mathcal{D}_{a2}$ are given, then a joint abstract domain $\mathcal{D}_a = \mathcal{D}_{a1} \cup \mathcal{D}_{a2}$ can be defined as follows: Let $\mathcal{D}_{a1} = (\mathcal{L}_{a1}, \mathcal{E}_{a1}, \mathcal{O}_{a1}, \mathcal{R}_{a1})$ and let $\mathcal{D}_{a2} = (\mathcal{L}_{a2}, \mathcal{E}_{a2}, \mathcal{O}_{a2}, \mathcal{R}_{a2})$. Then $\mathcal{D}_a = \mathcal{D}_{a1} \cup \mathcal{D}_{a2} = (\mathcal{L}_{a1} \cup \mathcal{L}_{a2}, \mathcal{E}_{a1} \cup \mathcal{E}_{a2}, \mathcal{O}_{a1} \cup \mathcal{O}_{a2}, \mathcal{R}_{a1} \cup \mathcal{R}_{a2})$. The joint abstract domain $\mathcal{D}_a$ fulfills the following property: if $C_a$ is an abstraction of $C_c$ with respect to $(\mathcal{D}_c, \mathcal{D}_{a1})$ or with respect to $(\mathcal{D}_c, \mathcal{D}_{a2})$, then $C_a$ is also an abstraction of $C_c$ with respect to $(\mathcal{D}_c, \mathcal{D}_a)$.*

**Proof:** The proof of this lemma is quite simple. If $C_a$ is an abstraction of $C_c$ with respect to $(\mathcal{D}_c, \mathcal{D}_{ai})$, then there exists a sequence abstraction mapping $\alpha$ and a sequence abstraction mapping $\beta$ as required in Definition 5. As it is easy to see, the same abstraction mappings will also lead to the respective case abstraction in $(\mathcal{D}_c, \mathcal{D}_a)$. □

**Lemma 7** (Multi-Level Hierarchy) *Let $(\mathcal{D}_0, \dots, \mathcal{D}_l)$ be an arbitrary multi-level hierarchy of domain descriptions. For the two-level description $(\mathcal{D}_c, \mathcal{D}_a)$ with $\mathcal{D}_a = \bigcup_{\nu=1}^{l} \mathcal{D}_\nu$ and $\mathcal{D}_c = \mathcal{D}_0$ holds that: if $C_a$ is an abstraction of $C_c$ with respect to $(\mathcal{D}_0, \dots, \mathcal{D}_l)$ then $C_a$ is also an abstraction of $C_c$ with respect to $(\mathcal{D}_c, \mathcal{D}_a)$.*

**Proof:** Let $C_\nu = \langle\langle s_0^\nu, s_m^\nu\rangle, \bar{o}^\nu\rangle$ be a case in domain $\mathcal{D}_\nu$ (intermediate state are denoted by $s_j^\nu$), let $C_0 = \langle\langle s_0^0, s_n^0\rangle, \bar{o}^0\rangle$ be a case in domain $\mathcal{D}_0$ (intermediate state are denoted by $s_i^0$), and let $C_\nu$ be an abstraction of case $C_0$ with respect to $(\mathcal{D}_0, \dots, \mathcal{D}_\nu)$. Then a sequence of cases $(\mathcal{C}_1, \dots, \mathcal{C}_{\nu-1})$ exists such that $\mathcal{C}_i$ is from the domain $\mathcal{D}_i$ and $\mathcal{C}_{i+1}$ is an abstraction of the case $\mathcal{C}_i$ with respect to $(\mathcal{D}_i, \mathcal{D}_{i+1})$ for all $i \in \{0, \dots, \nu-1\}$. Now we proof by induction over $\nu$ that $C_\nu$ is also an abstraction of $C_0$ with respect to $(\mathcal{D}_c, \mathcal{D}_a)$ (see figure 13). The basis ($\nu = 1$) is obvious: $C_1$ is abstraction of $C_0$ with respect to $(\mathcal{D}_0, \mathcal{D}_1)$ and is consequently also an abstraction with respect to $(\mathcal{D}_c, \mathcal{D}_a)$. Now, assume that the lemma holds for any cases up to the domain $\mathcal{D}_{\nu-1}$. It follows immediately that $C_{\nu-1}$ is an abstraction of $C_0$ with respect to $(\mathcal{D}_c, \mathcal{D}_a)$. Let $C_{\nu-1} = \langle\langle s_0', s_k'\rangle, \bar{o}'\rangle$ and let the intermediate states be denoted by $s_r'$. From Definition 5 follows, that a state abstraction mapping $\alpha$ and a sequence abstraction mapping $\beta$ exists, such that $\alpha(s_{\beta(r)}^c) = s_r'$ for all $r \in \{0, \dots, k\}$. Because $C_\nu$ is an abstraction of $C_{\nu-1}$ with respect to $(\mathcal{D}_{\nu-1}, \mathcal{D}_\nu)$, it also exists a state abstraction mapping $\alpha'$ and a sequence abstraction mapping $\beta'$ such that $\alpha'(s'_{\beta'(j)}) = s^\nu_j$ for all $j \in \{0, \ldots, m\}$. Now, we can define a state abstraction mapping $\alpha''(s) = \alpha'(\alpha(s))$ and a sequence abstraction mapping $\beta''(j) = \beta(\beta'(j))$. It is easy to see, that $\alpha''$ is a well defined state abstraction mapping $(s \supseteq s' \Rightarrow \alpha(s) \supseteq \alpha(s') \Rightarrow \alpha'(\alpha(s)) \supseteq \alpha'(\alpha(s')))$ and that $\beta''$ is a well defined sequence abstraction mapping $(\beta(\beta'(0)) = 0; \beta(\beta'(m)) = \beta(k) = n; u < v \Leftrightarrow \beta'(u) < \beta'(v) \Leftrightarrow \beta(\beta'(u)) < \beta(\beta'(v)))$. Furthermore it follows $\alpha''(s^c_{\beta''(j)}) = \alpha'(\alpha(s^c_{\beta(\beta'(j))})) = \alpha'(s'_{\beta'(j)}) = s^a_j$, leading to the conclusion that $C_\nu$ is an abstraction of $C_0$ with respect to $(\mathcal{D}_c, \mathcal{D}_a)$. $\square$

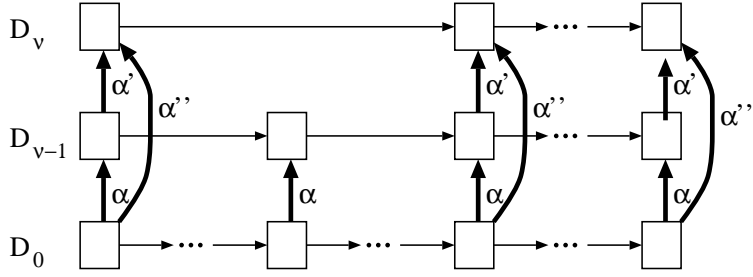

Figure 13: Abstraction mappings for hierarchies of abstraction spaces

**Theorem 8** (Correctness and completeness of the PABS algorithm) *If a complete SLD-refutation procedure is used in the PABS algorithm, then Case $C_a$ is an abstraction of case $C_c$ with respect to $(\mathcal{D}_c, \mathcal{D}_a)$ and the generic theory $\mathcal{A}$, if and only if $C_a = \text{PABS}(\langle \mathcal{D}_c, \mathcal{D}_a, \mathcal{A} \rangle, C_c)$.*

**Proof:**

*Correctness* ("⊆"): If $C_a$ is returned by PABS, then $\langle (o^a_1, \ldots, o^a_k), \alpha^{**}, \beta \rangle \in Paths$ holds [19] in phase-IV. We can define a state abstraction mapping $\alpha(s) := \{e \in \alpha^{**} | \mathcal{R}_c \cup \mathcal{A} \cup s \vdash e\}$, which, together with the sequence abstraction mapping $\beta$ will lead to the desired conclusion. For every operator $o^a_i$, we know by construction of phase-IV, that $\langle \beta(i-1), \beta(i), o^a_i, \tau \rangle \in G$ holds. By construction of phase-III, we can conclude that $s^a_{\beta(i-1)} \cup \mathcal{R}_a \vdash Pre_{o^a_i}$ holds and that consequently $\tau_\mathcal{E} \cup \mathcal{R}_a \vdash Pre_{o^a_i}$ also holds for the respective execution of the body of the while-loop in phase-IV. Since $\tau_\mathcal{E} \subseteq \alpha' \subseteq \alpha^{**}$ holds and $\vdash$ is a monotonic derivation operator, it is obvious that $\alpha(s^c_{\beta(i)}) \cup \mathcal{R}_a \vdash Pre_{o^a_i}$. Furthermore, the 'if for all'-test, which is executed before the extension of the path, ensures that $(s^a_{\beta(i-1)} \cap \alpha^{**}) \xrightarrow{o^a_i} (s^a_{\beta(i)} \cap \alpha^{**})$ holds. Together with the fulfillment of the precondition of the operator we have $\alpha(s^c_{\beta(i-1)}) \xrightarrow{o^a_i} \alpha(s^c_{\beta(i)})$. Thus, we have shown, $C_a$ is correct abstraction with respect to Definition 5.

*Completeness* ("⊇"): Assume, case $C_a = \langle \langle s^a_0, s^a_m \rangle, (o^a_1, \ldots, o^a_m) \rangle$ is an abstraction of $C_c$ based on a deductively justified state abstraction mapping. Then there exists a state ab-straction mapping $\alpha$ and a sequence abstraction mapping $\beta$ such that $\alpha(s^c_{\beta(i-1)}) \xrightarrow{o^a_i} \alpha(s^c_{\beta(i)})$ holds for all $i \in \{1, \ldots, m\}$. Since $\alpha$ is deductively justified by $\mathcal{A}$, it follows by construction of phase-II, that $\alpha(s^c_{i-1}) \subseteq s^a_{i-1}$. Since $\vdash$ is a monotonic derivation operator, the precondition of $o^a_i$ is also fulfilled in $s^a_{\beta(i-1)}$. Furthermore, the addlist of the operator is fulfilled in $\alpha(s^c_{\beta(i)})$ and is consequently also fulfilled in $s^a_i$. By the construction of phase-III, it is now guaranteed, that $\langle \beta(i-1), \beta(i), o^a_i, \tau \rangle \in G$. Now, we would like to show, that in phase-IV:

- there exists a sequence of assignments to the variable $Paths$, such that $\langle (), \beta_0, \alpha^*_0 \rangle \in Paths$, $\langle (o^a_1), \beta_1, \alpha^*_1 \rangle \in Paths$, $\ldots$, $\langle (o^a_1, \ldots, o^a_m), \beta_m, \alpha^*_m \rangle \in Paths$ ,

- $\beta_k(\nu) = \beta(\nu)$ for $\nu \in \{0, \ldots, k\}$

- $(\alpha^*_k \cap s^a_l) \subseteq \alpha(s^c_l)$ for $l \in \{1, \ldots, n\}$ and

- $\alpha^*_k \supseteq \bigcup_{l=1}^k Add_{o^a_l}$.

---

19. Note that $\alpha^{**}$ refers to the set finally constructed after termination of the while-loop. We use $\alpha^*$ to denote the respective set during the construction in this loop.

The proof is by induction on i. The induction basis is obvious due to the initialization of the $Paths$ variable. Now, assume that $\langle (o^a_1, \ldots, o^a_k), \beta_k, \alpha^*_k \rangle \in Paths$ (with $k < m$) at some state of the execution of phase-IV. Since, $\langle \beta(k), \beta(k+1), o^a_{k+1}, \tau \rangle \in G$ holds as argued before, and $\beta(k) = \beta_k(k)$ by induction hypothesis, the selected operator sequence is tried to be extended by $o^a = o^a_{k+1}$ in the body of the while-loop. Additionally, we know, that $\tau_\mathcal{E}$ contains exactly those sentences which are required to proof the precondition of $o^a_{k+1}$. Note, that since the SLD-resolution procedure is assumed to be complete and $o^a_{k+1}$ is applicable in $\alpha(s^c_k)$, $\tau_\mathcal{E}$ is required to proof the preconditition of $o^a$ if and only if $\tau_\mathcal{E} \subseteq \alpha(s^c_{\beta(k)})$. Since $\alpha$ is deductively justified, $\forall e \in \tau_\mathcal{E}, \forall l \in \{1, \ldots, m\}$ holds: $e \in \alpha(s^c_{\beta(l)})$ if $s^c_{\beta(l)} \cup \mathcal{R}_c \cup \mathcal{A} \vdash e$. By construction of the $s^a_l$, $\forall e \in \tau_\mathcal{E}, \forall l \in \{1, \ldots, m\}$ holds: $e \in \alpha(s^c_{\beta(l)})$ if $e \in s^a_l$. Consequently, $\tau_\mathcal{E} \cap s^a_l \subseteq \alpha(s^c_l)$ for all $l \in \{1, \ldots, m\}$. On the other hand, we also know that $o^a_{k+1}$ leads to $\alpha(s^c_{\beta(k+1)})$. Consequently, $Add_{o^a_{k+1}} \subseteq \alpha(s^c_{\beta(k+1)})$. Following the same argumentation as above, we can conclude that $(Add_{o^a_{k+1}} \cap s^a_l) \subseteq \alpha(s^c_l)$ for all $l \in \{1, \ldots, m\}$. Consequently, for $\alpha' = \alpha^*_k \cup \tau_\mathcal{E} \cup Add_{o^a_{k+1}}$ holds that $\alpha' \cap s^a_l \subseteq \alpha(s^c_l)$. Now, we can conclude that $Paths$ is extended by $o^a_{k+1}$ as follows. Since $\alpha(s^c_{\beta(\nu-1)}) \xrightarrow{o^a_\nu} \alpha(s^c_{\beta(\nu)})$ holds and that $Add_{o^a_\nu} \in \alpha'$ and $(\alpha' \cap s^a_{\beta(\nu)}) \subseteq \alpha(s^c_{\beta(\nu)})$, we can immediately follow that $(\alpha' \cap s^a_{\beta(\nu-1)}) \xrightarrow{o^a_\nu} (\alpha' \cap s^a_{\beta(\nu)})$. Consequently, $\langle (o^a_1, \ldots, o^a_k, o^a_{k+1}), \alpha^*_{k+1}, \beta_{k+1} \rangle \in Paths$ with $\alpha^*_{k+1} = \alpha'$ and $\beta_{k+1}(\nu) = \beta_k(\nu) = \beta(\nu)$ for $\nu \in \{1, \ldots, k\}$ and $\beta_{k+1}(k+1) = \beta(k)$. So, the induction hypothesis is fulfilled for $k+1$. Thereby, it is shown that $C_a$ is returned by Pabs. $\square$

## Acknowledgements

The authors want to thank Agnar Aamodt, Jaime Carbonell, Padraig Cunningham, Subbarao Kambhampati, Michael M. Richter, Manuela Veloso, as well as all members of our research group for many helpful discussions and for remarks on earlier versions of this paper. Particularly, we want to thank Padraig Cunningham for carefully proof-reading the recent version of the paper. We are also greatly indebted to the anonymous JAIR reviewers who helped to significantly improve the paper. This research was partially supported by the German "Sonderforschungsbereich" SFB-314 and the Commission of the European Communities (ESPRIT contract P6322, the INRECA project). The partners of INRECA are AcknoSoft (prime contractor, France), tecInno (Germany), Irish Medical Systems (Ireland) and the University of Kaiserslautern (Germany).